\documentclass{article}

\usepackage[accepted]{icml2021}
\usepackage{commands}

\begin{document}


\twocolumn[
\icmltitle{Generalization Bounds in the Presence of Outliers: a Median-of-Means Study}




\begin{center}
\textbf{Pierre Laforgue$^1$}\hspace{2.3cm}
\textbf{Guillaume Staerman$^2$}\hspace{2.3cm}
\textbf{Stephan Cl\'emen\c{c}on$^2$}\\\vskip 0.2cm
$^1$ DSRC \& Dept. of Computer Science, Universit\`{a} degli Studi di Milano, Italy\\\vskip 0.06cm
$^2$ LTCI, T\'el\'ecom Paris, Institut Polytechnique de Paris, France\\\vskip 0.15cm
\textit{Correspondence to:} \href{mailto:pierre.laforgue@unimi.it}{pierre.laforgue@unimi.it}\vskip 0.67cm
\end{center}


]

\hypersetup{urlcolor={blue}}




\begin{abstract}
In contrast to the empirical mean, the Median-of-Means (MoM) is an estimator of the mean $\theta$ of a square integrable r.v. $Z$, around which accurate nonasymptotic confidence bounds can be built, even when $Z$ does not exhibit a sub-Gaussian tail behavior.
Thanks to the high confidence it achieves on heavy-tailed data, MoM has found various applications in machine learning, where it is used to design training procedures that are not sensitive to atypical observations.
More recently, a new line of work is now trying to characterize and leverage MoM's ability to deal with corrupted data.
In this context, the present work proposes a general study of MoM's concentration properties under the contamination regime, that provides a clear understanding of the impact of the outlier proportion and the number of blocks chosen.
The analysis is extended to (multisample) $U$-statistics, \textit{i.e.} averages over tuples of observations, that raise additional challenges due to the dependence induced.
Finally, we show that the latter bounds can be used in a straightforward fashion to derive generalization guarantees for pairwise learning in a contaminated setting, and propose an algorithm to compute provably reliable decision functions.
%
%
%
%
\end{abstract}

\section{Introduction}

There are undoubtedly two major reasons for the success of modern machine learning techniques: on the one hand, the increasing availability of massive datasets, on the other, the existence of computationally efficient and statistically accurate estimation procedures.
If the constant improvement of data acquisition technologies, such as the Internet of Things (IoT), enables today to collect considerable datasets in an automatic fashion, it also raises numerous challenges on the estimation side, due to the heterogeneity and possible corruption of the observations acquired.
From a statistical perspective, two frameworks have been introduced to model these aspects: (1) the heavy-tailed framework, where only low-order moments are assumed to be finite for the data distribution, (2) the $\varepsilon$-contamination model \citep{huber1964}, where the available dataset is supposed to be corrupted by a proportion $\varepsilon$ of outliers.

Univariate mean estimation plays a critical role in many statistical learning problems, ranging from classification and regression to ranking or generative modeling.
Although the empirical mean appears as a natural candidate, it has been unfortunately shown to dramatically fail under either of the two models discussed above.
Consider a sample $\mathcal{S}_n = \{Z_1, \ldots, Z_n\}$ composed of $n$ independent identically distributed (i.i.d.) realizations of the real-valued random variable $Z$, with distribution $P$.
It is well known that for the empirical mean $\hat{\theta} = (1/n)\sum_{i=1}^n Z_i$ to exhibit a sub-Gaussian tail behavior, it is required that distribution $P$ must also be \mbox{sub-Gaussian}, \textit{i.e.} there exists $\rho > 0$ such that $\mathbb{E}_P[e^{\lambda Z}]\leq e^{\lambda^2\rho^2/2}$ for all $\lambda\in \mathbb{R}$.
In contrast, in the heavy-tailed model, one is rather interested by estimates enjoying similar guarantees but under much weaker assumptions, such as having only a finite variance, see the following assumption, supposed to be verified throughout this paper.
\begin{assumption}\label{hyp:variance}
There exist $\theta$ and $\sigma^2 < +\infty$ such that $
\mathbb{E}_P\left[Z\right] = \theta$, and $\mathrm{Var}_P(Z) = \sigma^2$.
\end{assumption}

\begin{figure}[!t]
\begin{center}
\includegraphics[width=\columnwidth, page=1]{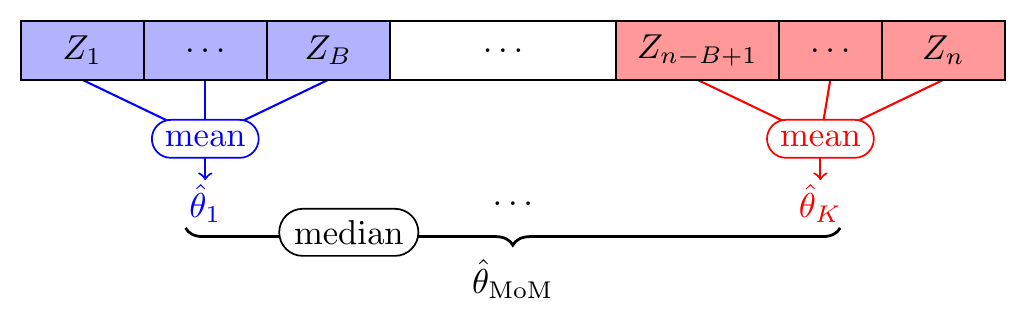}
\caption{The MoM estimator.}
\label{fig:mom}
\end{center}
\vspace{-0.3cm}
\end{figure}

The Median-of-Means (MoM) is one of the mean estimators that achieve a sub-Gaussian behavior under \Cref{hyp:variance}.
Independently introduced during the 1980s \citep{nemirovsky1983problem,jerrum1986random}, the Median-of-Means is a mean estimator that is easy to compute, while exhibiting attractive robustness properties.
For a predefined level of confidence $1-\delta$, with $\delta\in [e^{1-n/2}, 1[$, the MoM estimator is built as follows.
Set $K = \lceil \log(1/\delta) \rceil\leq n$, denoting by $x\in \mathbb{R}\mapsto \lceil x \rceil$ the ceiling function, and partition sample $\mathcal{S}_n$ into $K$ disjoint blocks $\mathcal{B}_1,\; \ldots,\; \mathcal{B}_K$ of size $B = \lfloor n / K \rfloor$, denoting by $x\in \mathbb{R}\mapsto \lfloor x \rfloor$ the floor function.
For $k \le K$, compute the empirical mean based on block $\mathcal{B}_k$: $\hat{\theta}_k = (1/B) \sum_{i \in \mathcal{B}_k} Z_i$.
The Median-of-Means $\hat{\theta}_\mathrm{MoM}$ is finally obtained by computing the median of the  block averages (see also \Cref{fig:mom}):
\begin{equation}\label{eq:basic_MoM}
\hat{\theta}_\mathrm{MoM} = \mathrm{median}(\hat{\theta}_1, \ldots, \hat{\theta}_K).
\end{equation}

The recent resurgence of interest for MoM in the statistical literature dates back to the seminal deviation studies by \citet{audibert2011robust} and \citet{catoni2012challenging}, that propose to assess an estimator through its deviation probabilities, rather than by computing its quadratic risk.
Extensively studied since then, MoM now benefits from a large corpus of concentration results.
For instance, a proof of its behavior under \Cref{hyp:variance} can be found in \citet{devroye2016sub}.
\begin{proposition}\label{prop:standard_mom}
\citep{devroye2016sub} Suppose that an i.i.d. sample $\mathcal{S}_n$ is drawn from $P$, satisfying \Cref{hyp:variance}.
Then, for any $\delta \in [e^{1-n/2}, 1[$, choosing $K = \lceil \log(1/\delta)\rceil$, it holds with probability at least $1- \delta$:
\begin{equation}\label{eq:mom_variance}
\big|\hat{\theta}_\mathrm{MoM} - \theta\big| ~\le~ 2\sqrt{2}e~\sigma\sqrt{\frac{1 + \log(1/\delta)}{n}}.
\end{equation}
\end{proposition}

These concentration results have further been extended to random vectors, through different generalizations of the median in a multidimensional setting \citep{minsker2015geometric,hsu2016loss,lugosi2017sub}, and to $U$-statistics (\citet{joly2016robust} for the degenerate case, \citet{laforgue2019medians} with randomized blocks) among other extensions.
Such interesting properties in the presence of heavy-tailed data has given birth to numerous applications in statistical learning.
This includes \textit{e.g.} an adaptation of the Upper Confidence Bound (UCB) bandit algorithm in \citet{bubeck2013bandits}, of Empirical Risk Minimization (ERM) in \citet{brownlees2015empirical}, or the more general framework of MoM-tournaments \citep{lugosi2019risk} and Le Cam's approach \citep{lecue2019learning}.

A recent line of work is now trying to change perspective, abandoning the heavy-tailed framework to focus on MoM's behavior within the Huber's contamination model.
Formally, the assumption considered in this paper is as follows.
%
\begin{assumption}\label{hyp:framework}
The sample $\mathcal{S}_n=\{Z_1,\; \ldots,\; Z_n\}$ contains $n - n_\mathsf{O}$ \emph{inliers} drawn i.i.d. according to distribution $P$, and $n_\mathsf{O}$ \emph{outliers}, upon which no assumption is made.
We denote by $\varepsilon = n_\mathsf{O}/n$ the fraction of outliers among sample $\mathcal{S}_n$.
\end{assumption}

\begin{remark}
We stress that \Cref{hyp:framework} can be related to the standard Huber's contamination model, which assumes that $\mathcal{S}_n$ is drawn i.i.d. from the mixture $\widetilde{P} = (1 - \zeta)P + \zeta A$, where $\zeta\in (0,1)$ and $A$ is an arbitrary distribution. Working under \Cref{hyp:framework} simply means working under this model, conditioned upon the event that the (random) number of observations actually generated by $A$ is equal to $n_\mathsf{O}$, whose marginal is a Binomial law of size $n$ and parameter $\zeta$.
\end{remark}

\Cref{hyp:framework} is thus addressed through the general angle of MoM-minimization in \citet{lecue2018robust}, while \citet{monk} develops an application to Maximum Mean Discrepancy and outlier-robust mean embedding.
\citet{depersin2019} proposes a sub-Gaussian MoM-inspired multidimensional estimator computable in almost linear time, and \citet{depersin2020robust} studies a multivariate estimator based on one-dimensional projections.
However, all these works rely on \textit{ad-hoc} assumptions that are quite difficult to interpret.
For instance, \citet{lecue2018robust} uses unusual outlier-adapted Rademacher complexities, while the choice of $K$ is based on unknown constants in \citet{depersin2020robust}, or defined implicitly in \citet{monk}.
In \citet{depersin2019}, the choice of $K$ incidentally reduces the analysis to the case where $\varepsilon \le 0.33\%$.

In contrast, this paper proposes a unified and insightful study of the concentration properties of (univariate) MoM-based estimators under the contamination regime of \Cref{hyp:framework}.
In particular, we show that MoM is able to handle up to $50\%$ of outliers, at the price of a degraded constant though.
Indeed, our bounds allow to encapsulate the impact of the proportion of outliers $\varepsilon$ into constant terms only.
As this performance can be achieved through a multitude of values for the number of blocks $K$, we also fully characterize the impact of this choice, exemplified by $4$ representative strategies.
Another important insight given by our analysis is that MoM may handle both outliers and heavy-tailed inliers, but on limited range of confidence levels only.
Assuming instead the inliers to be sub-Gaussian, we show that MoM becomes efficient on a wide interval, allowing next to derive bounds in expectation (we are not aware of similar results for MoM) under the following assumption stipulating that the number of outliers $n_\mathsf{O}$ grows sub-linearly with $n$.
\begin{assumption}\label{ass:sub_linear}
There exist constants $C_\mathsf{O} \ge 1$ and $\alpha_\mathsf{O} \in [0, 1[$ such that: $\forall n\geq 1$, $n_\mathsf{O} \le C_\mathsf{O}^2~n^{\alpha_\mathsf{O}}$.
\end{assumption}
The extension to multisample $U$-statistics raises interesting discussions about the fractions of outliers authorized by the different approaches.
We then show that our bounds can be easily combined with standard class complexities (VC-dimension, entropy) to produce generalization bounds for pairwise learning in the presence of outliers.
We finally detail an algorithm whose outputs satisfy these guarantees.

The rest of the article is organized as follows.
In \Cref{sec:revisit} are stated the concentration results for the MoM estimator and its extensions to (multisample) $U$-statistics under the regime of \cref{hyp:framework}.
The applications to learning theory are detailed in \Cref{sec:learning}.
Due to space constraints, technical proofs, as well as numerical results validating our theoretical findings, are deferred to the Supplementary Material.

\textbf{Related Works.}
Of course, the Median-of-Means is not the sole estimator to achieve sub-Gaussian behavior under the contaminated model.
One may for instance mention the trimmed mean \citep{oliveira2019, lugosi2020}.
The existing bounds however exhibit a complex dependence with respect to $\varepsilon$, in contrast to our results.
One of the important drawback of MoM lies in its computational intractability in high dimension, motivating an important line of research in the field of robust mean estimation \citep{diakonikolas2016,lai2016,cheng2018highdimensional,hopkins2018sub,cherapanamjeri2019,prasad2019,prasad2020}.
Our analysis essentially differs from these works in three ways: $(i)$ as we ultimately target to derive learning bounds, \textit{i.e.} bounds on risk estimates, we shall focus on univariate estimators, bypassing also the computational difficulties with multidimensional MoMs, $(ii)$ it allows for a complete characterization of the impact of $\varepsilon$ and $K$, and $(iii)$ the extension to $U$-statistics is entirely new to the best of our knowledge.


\section{Concentration of MoM-based Estimators in the Presence of Outliers}
\label{sec:revisit}

In this section, we study the concentration properties of MoM, and those of its recent extensions to $U$-statistics, under the contamination regime of \Cref{hyp:framework}.


\subsection{Concentration Bounds for MoM}

In this section, we prove an extension of bound \eqref{eq:mom_variance} when the sample $\mathcal{S}_n$ is corrupted according to \Cref{hyp:framework}.
As revealed by \Cref{prop:standard_mom}, when $\mathcal{S}_n$ is not corrupted, $K$ must be set depending on the targeted confidence $\delta$.
When outliers are added, $K$ must also be chosen according to the outlier ratio~$\varepsilon$.
Roughly, we want $K > 2n_\mathsf{O}$ to ensure that blocks without outliers are in majority.
However, if $K$ is too large MoM tends to the median, which is a bad estimator of the mean in general.
To correctly calibrate $K$, we introduce a mapping $\alpha\colon[0, 1/2] \rightarrow [0, 1]$ upper bounding $\varepsilon \mapsto 2\varepsilon$.
This way, setting $K \approx \alpha(\varepsilon)n > 2\varepsilon n = 2n_\mathsf{O}$ satisfies the outlier constraint, while refraining from choosing too large values if the bound $\alpha$ is tight enough.
Based on $\alpha$, we derive functions $\beta, \gamma, \Gamma, \Delta$, that appear through the computations and shape the bounds established in \Cref{prop:mom}.
\begin{assumption}\label{hyp:config}
The mapping $\alpha\colon [0, 1/2] \rightarrow [0, 1]$ satisfies
\begin{equation*}
\forall \varepsilon \in ]0, 1/2[, \quad 2\varepsilon < \alpha(\varepsilon) < 1.
\end{equation*}
From mapping $\alpha$, we define the following functions:
\begin{align*}
&\beta\colon\varepsilon \mapsto \frac{2\alpha(\varepsilon)}{\alpha(\varepsilon) - 2\varepsilon}, \qquad
&&\gamma\colon \varepsilon \mapsto \frac{\sqrt{\alpha(\varepsilon)}(\alpha(\varepsilon) - \varepsilon)}{(\alpha(\varepsilon) - 2\varepsilon)^{\frac{3}{2}}},\\
&\Gamma\colon \varepsilon \mapsto \sqrt{\frac{\alpha(\varepsilon)}{\alpha(\varepsilon)-2\varepsilon}},
&&\Delta\colon \varepsilon \mapsto \sqrt{\frac{\alpha(\varepsilon)}{\varepsilon}}.
\end{align*}
\end{assumption}

\begin{figure*}[!h]
\begin{center}
\begin{subfigure}[b]{0.24\textwidth}
\includegraphics[width=\textwidth]{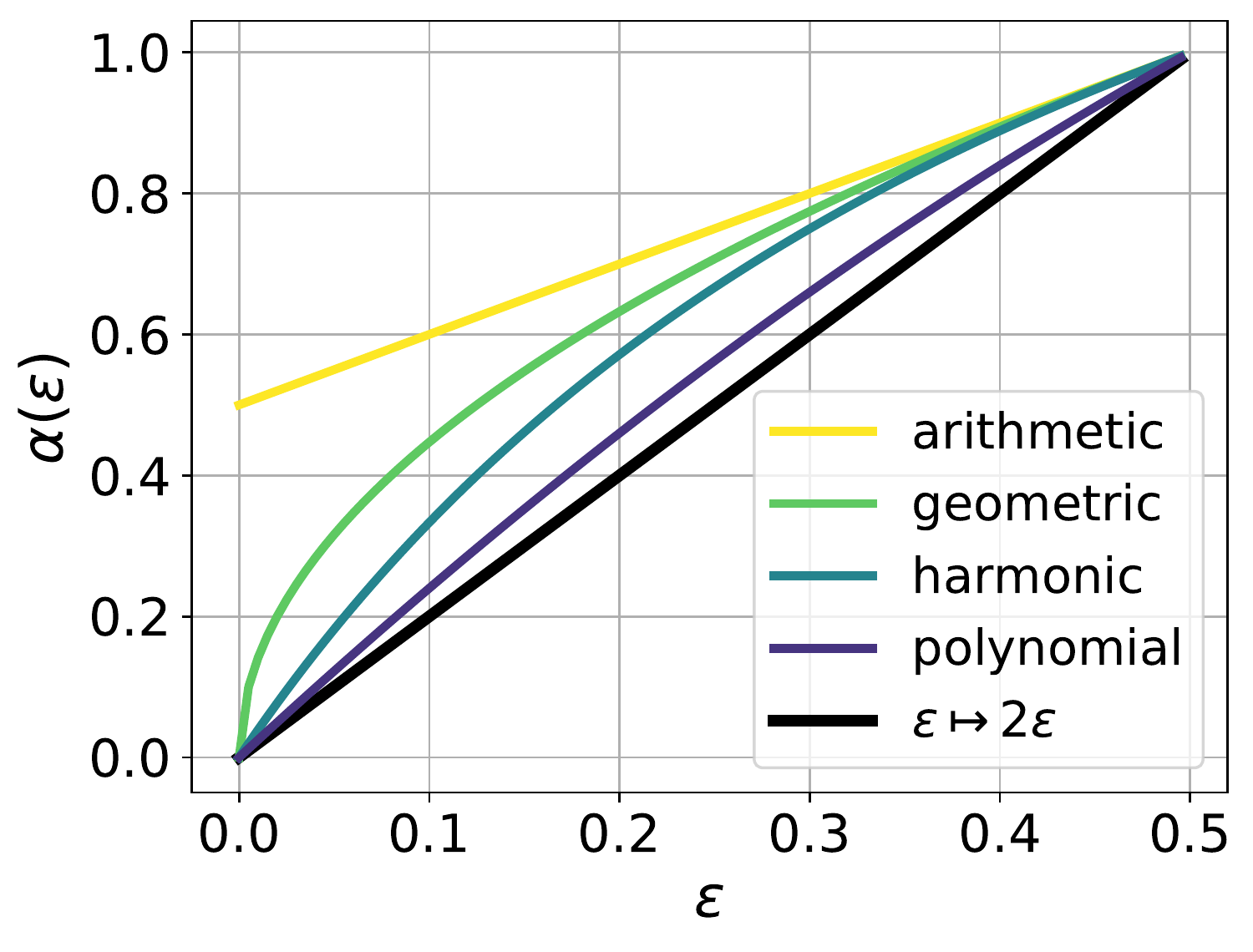}
\vspace{-0.4cm}
\caption{Upper bounds $\alpha(\varepsilon)$}
\label{fig:alphas}
\end{subfigure}
\hfill
\begin{subfigure}[b]{0.24\textwidth}
\includegraphics[width=\textwidth]{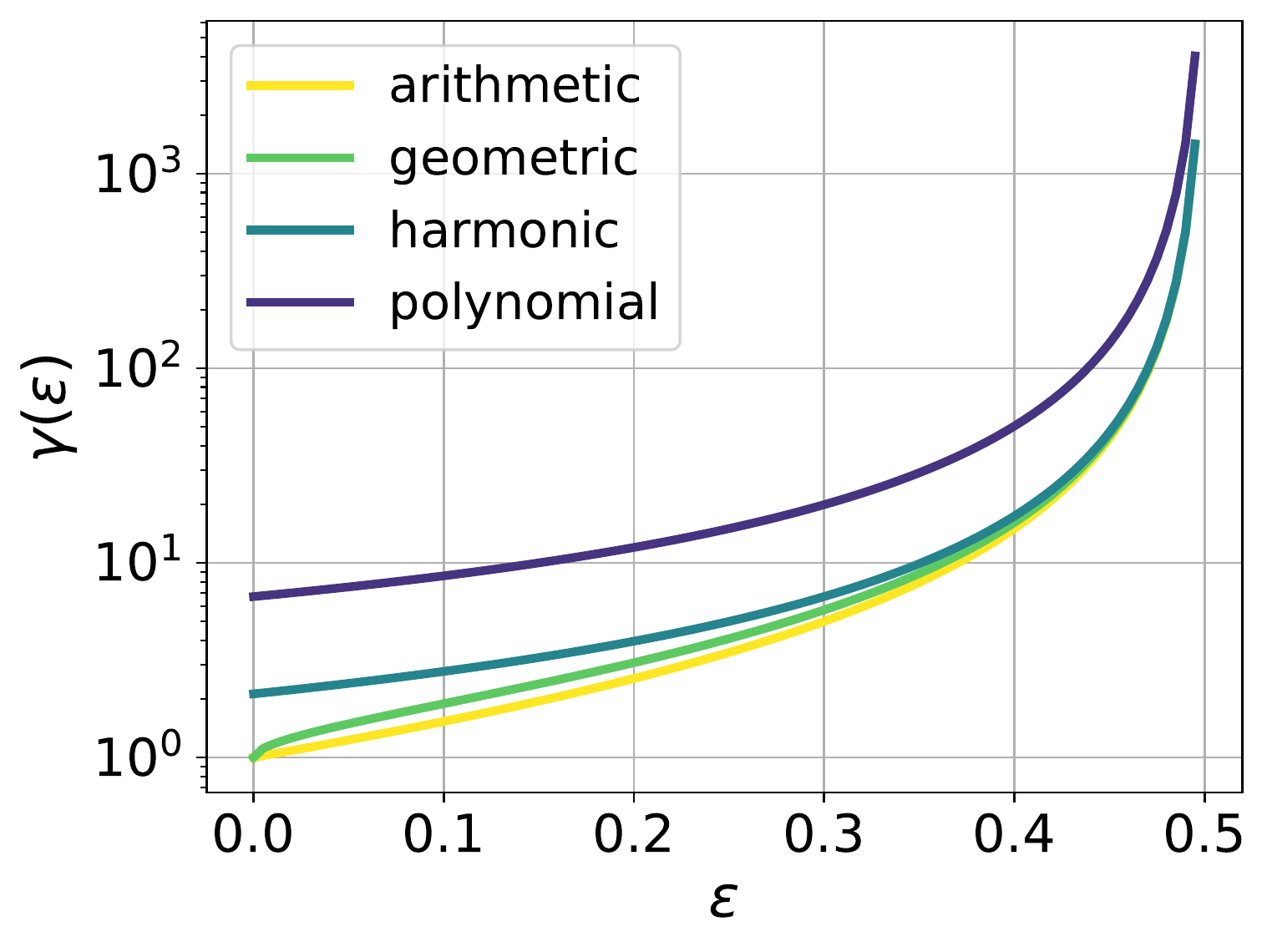}
\vspace{-0.4cm}
\caption{Constants $\gamma(\varepsilon)$}
\label{fig:gammas}
\end{subfigure}
\hfill
\begin{subfigure}[b]{0.24\textwidth}
\includegraphics[width=\textwidth]{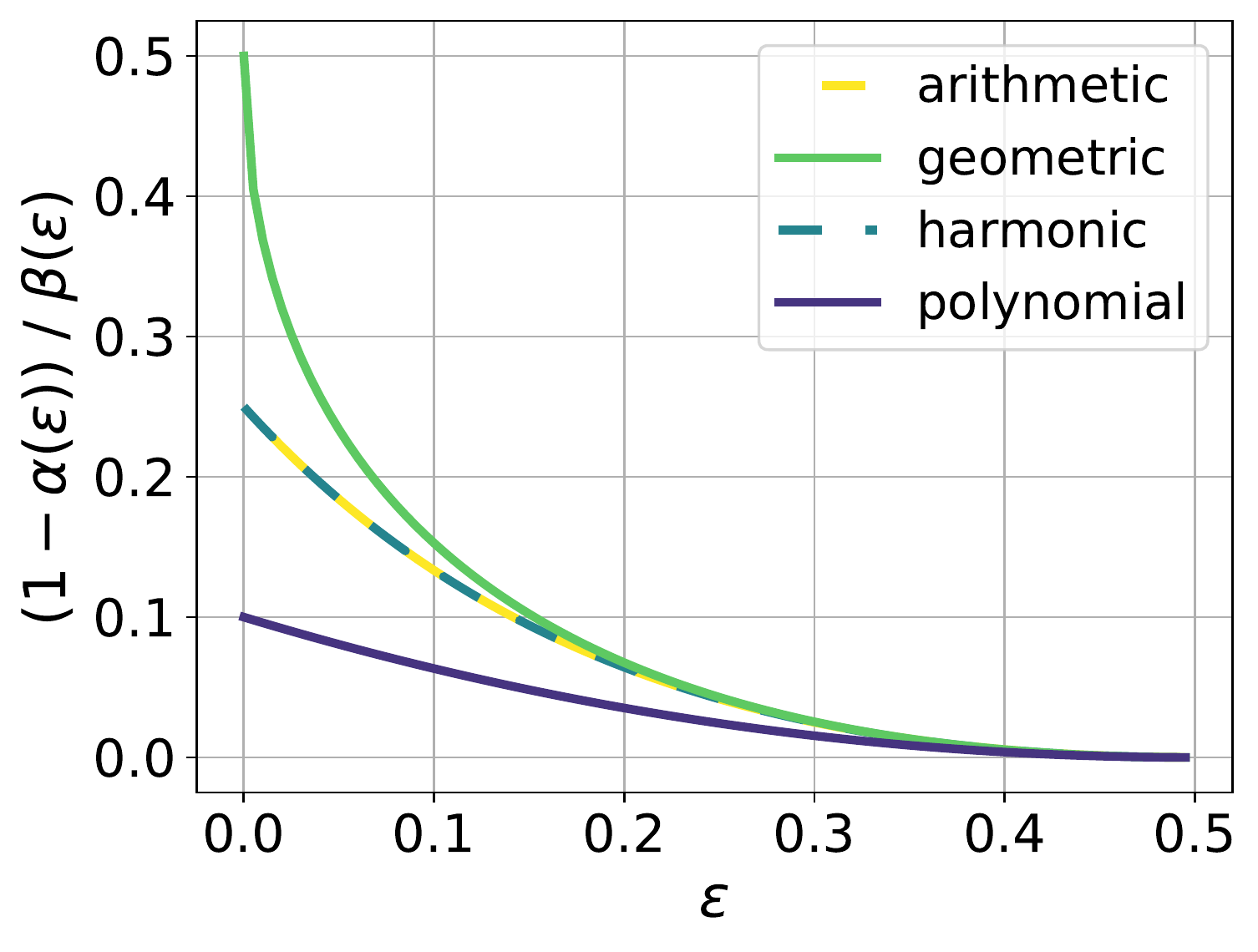}
\vspace{-0.4cm}
\caption{Range sizes $s(\varepsilon)$ in log scale}
\label{fig:ranges}
\end{subfigure}
\hfill
\begin{subfigure}[b]{0.24\textwidth}
\includegraphics[width=\textwidth]{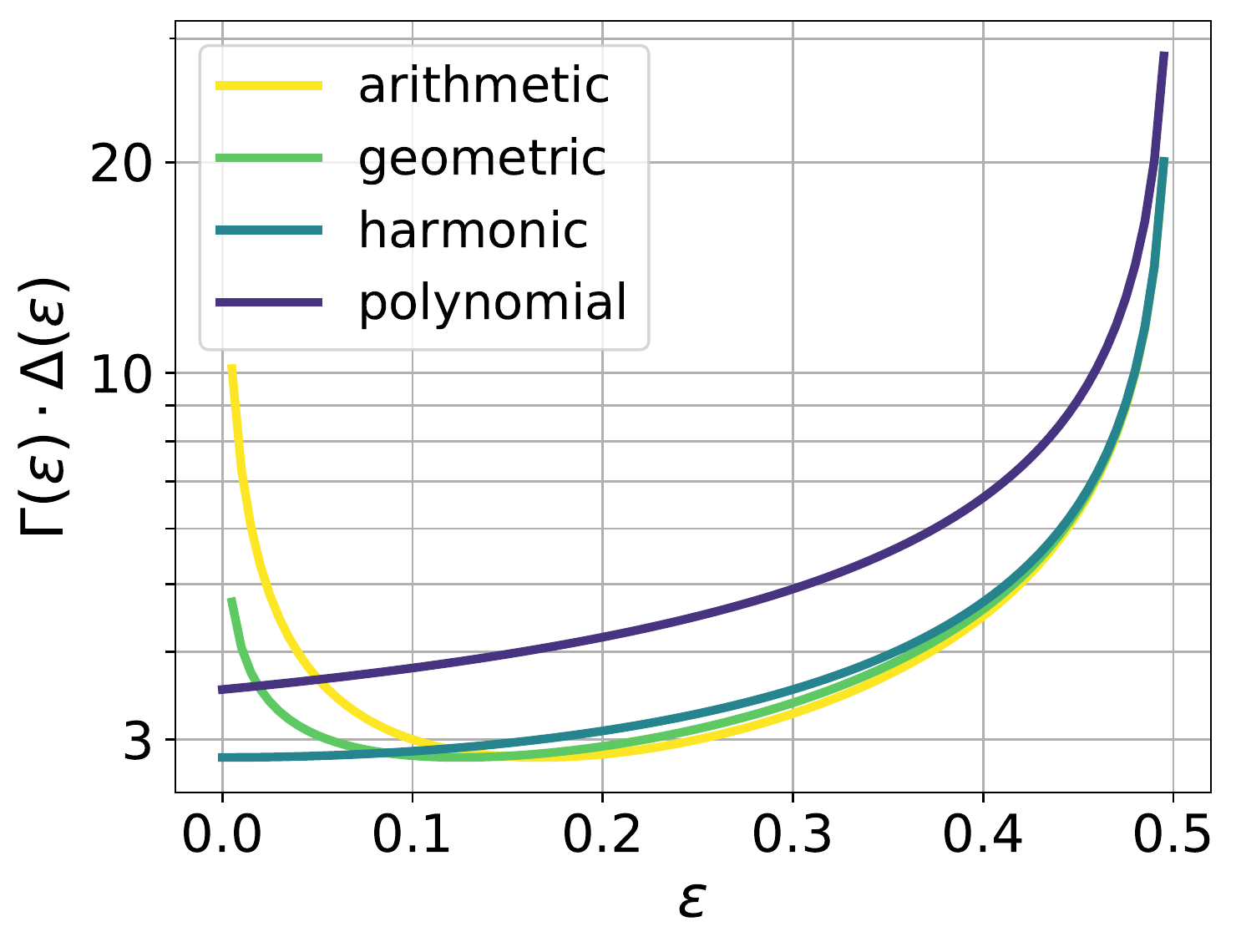}
\vspace{-0.4cm}
\caption{Constants $\Gamma(\varepsilon) \cdot \Delta(\varepsilon)$}
\label{fig:Gamma_Deltas}
\end{subfigure}
\caption{Influence of the chosen mapping $\alpha$ on the constants.}
\label{fig:influence}
\end{center}
\vspace{-0.3cm}
\end{figure*}

We now give several examples of mappings $\alpha$ satisfying \Cref{hyp:config}.
Their plots can be found in \Cref{fig:alphas}.
The reader is referred to \Cref{apx:plots} (\Cref{tab:summary}, \Cref{fig:influence_apx}) for details about the corresponding functions $\beta, \gamma, \Gamma, \Delta$.
%

\begin{example}\label{example}
As we want $2\varepsilon < \alpha(\varepsilon) < 1$, natural choices for $\alpha$ involve the means of $2\varepsilon$ and $1$, taken either arithmetic, geometric or harmonic.
The last example is a polynomial.
\begin{table}[!ht]
\begin{center}
\begin{tabular}{ccccc}\toprule
& {\sc\scriptsize Arithmetic} & {\sc\scriptsize Geometric} & {\sc\scriptsize Harmonic} & {\sc\scriptsize Polynomial}\\\midrule
$\alpha(\varepsilon)$ & $\displaystyle\frac{1 + 2\varepsilon}{2}$ & $\displaystyle\sqrt{2\varepsilon}$ & $\displaystyle\frac{4\varepsilon}{1 + 2\varepsilon}$ & $\displaystyle\varepsilon\Big(\frac{5}{2} - \varepsilon\Big)$\\\bottomrule
\end{tabular}
\end{center}
\end{table}
\end{example}

The next proposition describes the concentration of MoM under the contamination regime of \Cref{hyp:framework}.
\begin{proposition}\label{prop:mom}
Suppose that sample $\mathcal{S}_n$ and mapping $\alpha$ satisfy \Cref{hyp:framework,hyp:config} respectively.
Define functions $\beta, \gamma, \Gamma, \Delta$ according to \Cref{hyp:config}.
Then, for any $\delta \in [e^{-n/\beta(\varepsilon)}, e^{-n\alpha(\varepsilon)/\beta(\varepsilon)}]$, choosing $K = \left\lceil \beta(\varepsilon)\log(1/\delta)\right\rceil$, it holds with probability at least $1-\delta$:
\begin{equation}\label{eq:bound1}
\big|\hat{\theta}_\mathrm{MoM} - \theta\big| ~\le~ 4\sqrt{e}\sigma~\gamma(\varepsilon)~\sqrt{\frac{1 + \log(1/\delta)}{n}}.
\end{equation}

If in addition distribution $P$ is $\rho$ sub-Gaussian, then for all $\delta \in ]0, e^{-4n\alpha(\varepsilon)}]$, with $K = \lceil \alpha(\varepsilon) n \rceil$, it holds w.p.a.l. $1 - \delta$:
\begin{equation}\label{eq:bound2}
\big|\hat{\theta}_\mathrm{MoM} - \theta\big| ~\le~ 4\rho~\Gamma(\varepsilon)~\sqrt{\frac{\log(1/\delta)}{n}}.
\end{equation}

If furthermore $n_\mathsf{O}$ satisfies \Cref{ass:sub_linear}, the same $K$ gives:
\begin{equation*}
\mathbb{E}\left[ \big|\hat{\theta}_\mathrm{MoM} - \theta\big|\right] ~\le~ 2\rho~\Gamma(\varepsilon)\left(4C_\mathsf{O}~\frac{\Delta(\varepsilon)}{n^{(1 - \alpha_\mathsf{O})/2}} + \sqrt{\frac{\pi}{n}}\right).
\end{equation*}
\end{proposition}

The technical proof is given in \Cref{apx:proof_mom}.
Its argument essentially consists in using that the MoM estimator \eqref{eq:basic_MoM} has a similar behavior to that of a majority of block means.
The condition $K > 2n_\mathsf{O}$ is strengthened into $K \ge \alpha(\varepsilon)n$, where the function $\alpha$ is a strict upper bound of the mapping $\varepsilon \mapsto 2\varepsilon$ on $]0, 1/2[$, ensuring that a fraction $\eta(\varepsilon) = (\alpha(\varepsilon) - \varepsilon)/\alpha(\varepsilon)>1/2$ of ``sane'' blocks (\textit{i.e.} including none of the $n_\mathsf{O}$ outliers) actually constitutes a majority of blocks.
One may then focus on the sane blocks deviations only, which is controlled by means of the concentration properties of a Binomial random variable.
The sub-Gaussian assumption allows for a sharper analysis of what happens on the sane blocks, resulting in an improved confidence interval (notice that the choice of $K$ then becomes independent from $\delta$).
The expectation bound is finally obtained by integrating the tail probability bound derived in \Cref{eq:bound2}.
%


As revealed by \Cref{prop:mom}, the choice of $\alpha$ shapes the constant terms in the upper bounds, as well as the range of confidence levels for which they hold true (however, it does not affect the rate).
This subtle balance calls for in depth discussions to determine the optimal mapping $\alpha$.

{\bf A $\delta$-limited sub-Gaussian tail bound.}
We first point out that the main price to pay for extending the sub-Gaussian tail behavior of MoM to the contaminated framework of \Cref{hyp:framework} is the limited range of acceptable confidence levels $1-\delta$.
This type of limitation is typical of MoM's concentration results.
The lower limit value for $\delta$ is due to the constraint $K \le n$, and is not very compelling in practice as it decays to zero exponentially fast as $n$ increases.
The upper limit value comes from the constraint $ 2n_\mathsf{O} < K$ (or $\alpha(\varepsilon)n \le K$), and is specific to the contaminated framework.
It should be noticed that this restriction vanishes (\textit{i.e.} the upper limit value is $1$) when $\varepsilon=0$ for all mappings $\alpha$ given in \Cref{example}, except for  the arithmetic mean.
Observe also that the lower limit restriction is removed when assuming that $P$ is sub-Gaussian.
We incidentally underline that this assumption only applies to $P$, and not to $A$, so that any hope of using reliably the empirical mean remains vain.

{\bf About the constants.}
An interesting property of the bounds derived in \Cref{prop:mom} is that they fully encapsulate the impact of the proportion of outliers $\varepsilon$ into the constants $\gamma(\varepsilon)$ and $\Gamma(\varepsilon)$.
Naturally, the latter increase with $\varepsilon$, and tend to infinity as $\varepsilon$ goes to $1/2$, see \Cref{fig:gammas}.
This dependence w.r.t. $\varepsilon$ can be further explicited, as one may notice that there exist universal constants $c$ and $C$ such that for all mappings presented in \Cref{example}, it holds $\gamma(\varepsilon) \le c/(1 - 2\varepsilon)^{3/2}$ and $\Gamma(\varepsilon) \le C/\sqrt{1 - 2\varepsilon}$, see \Cref{tab:summary} for details.

{\bf Accuracy vs range of confidence levels.}
As previously mentioned, the choice of mapping $\alpha$ determines at the same time the range $[\exp(-n/\beta(\varepsilon)), \exp(-n\alpha(\varepsilon)/\beta(\varepsilon))]$ for which \Cref{eq:bound1} holds true with probability at least $1-\delta$, and the constant $\gamma(\varepsilon)$.
When $\varepsilon\in [0,1/2[$ is fixed, the quantity $\gamma(\varepsilon)$ monotonically decreases as $\alpha(\varepsilon)$ increases.
Indeed, one may easily check that it holds $(\partial \gamma^2_{\varepsilon}/\partial \alpha)(\alpha)=-4\varepsilon(\alpha-\varepsilon)^2/(\alpha-2\varepsilon)^4<0$, with the notation $\gamma^{2}_{\varepsilon}(\alpha)=\alpha(\varepsilon)(\alpha(\varepsilon) - \varepsilon)^2 / (\alpha(\varepsilon) - 2\varepsilon)^{3}$.
Hence, the larger $\alpha(\varepsilon)$, the smaller the constant in the upper bound, encouraging the practitioner to choose the arithmetic upper bound, see \Cref{fig:gammas}.
However, the choice of $\alpha$ also impacts the confidence range, mitigating this incentive.
Precisely, when $\varepsilon \in ]0, 1/2[$ is fixed, its size $s(\varepsilon)$ increases with $\alpha(\varepsilon)$ on $]2\varepsilon, \sqrt{2\varepsilon}]$, and decreases on $[\sqrt{2\varepsilon}, 1]$.
Indeed, at the log scale, it is equal to $s_{\varepsilon}(\alpha)=n(\alpha-2\varepsilon)(1-\alpha)/(2/\alpha)$, and $(\partial s_{\varepsilon}/\partial \alpha)(\alpha)=n(2\varepsilon-\alpha^2)/(2/\alpha^2) $ for $\alpha\in]0,\; 1/2[$.
As a consequence, starting from $\alpha(\varepsilon) = \sqrt{2\varepsilon}$ (\textit{i.e.} the geometric mean), increasing $\alpha(\varepsilon)$ indeed reduces $\gamma(\varepsilon)$, but at the price of a smaller range of the confidence levels, see \Cref{fig:ranges}.
A similar phenomenon occurs for the bound \eqref{eq:bound2}: there is a trade-off between the size of the range for the confidence levels and the order of magnitude of the constant $\Gamma(\varepsilon)$, both decreasing with $\alpha(\varepsilon)$.
After integration, this tradeoff can be seen in the opposition between constants $\Gamma(\varepsilon)$ and $\Delta(\varepsilon)$, which have inverse monotonicity w.r.t. $\alpha(\varepsilon)$, see \Cref{fig:Gamma_Deltas} for plots of their product.
The fact that $\Delta(\varepsilon) \rightarrow \infty$ when $\varepsilon \rightarrow 0$ for some choices of $\alpha$ may reflect an artifact of the proof technique.
Indeed, if $\varepsilon = n_\mathsf{O} = 0$, it is not allowed to multiply/divide by $\varepsilon$ in \Cref{eq:t_sup}.
In contrast, one may use $\delta \le 1/e$ instead of \Cref{eq:delta_inf}, which then gives a $1/\sqrt{n}$ term, with no dependence with respect to $\Delta$.

{\bf Rate bound.}
We underline that the rate $1/\sqrt{n^{1-\alpha_\mathsf{O}}}$ for the mean deviation is in accordance with the expectations.
Indeed, MoM trades the ability of discarding outliers for the degradation of its statistical guarantees to those of one single sane block, of order $1/\sqrt{B} \sim \sqrt{K/n}\sim \sqrt{n_\mathsf{O}/n}$, as $K$ is roughly of the order of $n_\mathsf{O}$.
Hence, if $n_\mathsf{O}$ grows linearly with~$n$, then $B$ stays bounded and guarantees do not improve with $n$.
This also highlights the importance of not choosing a too rough upper bound $\alpha$.
We finally highlight that this rate is optimal.
Indeed, our bounds are obtained after conditioning upon the observations and, as can be seen by examining proofs, they cannot be refined, insofar as they simply rely on exact computations of the binomial distribution.
%


{\bf Unknown $\varepsilon$.}
In practice, the proportion of outliers $\varepsilon$ is generally unknown, preventing from using it to calibrate $K$.
We emphasize that the above stated bounds may still be used with an overestimation of $\varepsilon$, at the price of a deterioration of $\gamma(\varepsilon), \Gamma(\varepsilon)$ and $s(\varepsilon)$ though.
%


{\bf Related work.}
Although they are quite similar in spirit, six critical points distinguish \Cref{prop:mom} from Theorem 1 in \citet{monk}.
(1) It is important to notice first that \Cref{prop:mom} focuses on the deviations of scalar MoMs, while Theorem 1 in \citet{monk} addresses that of particular kernel mean embeddings, defined as MoM minimizers.
(2) This being said, our choice of $K$ can be computed explicitly from the total proportion of outliers $\varepsilon$, and the targeted confidence $\delta$.
In contrast, the number of blocks in \citet{monk} depends on the proportion of outliers with respect to the number of blocks itself, resulting in a recursive definition, hard to disambiguate.
This inherent difficulty is typically overcome here by reparameterizing using $\eta(\varepsilon)$ .
(3) As a consequence, our bound features the true and fixed proportion of outliers $\varepsilon$ within the sample, while \citet{monk} use the proportion w.r.t. the number of blocks, that may change with it.
(4) Additionally, their range of admissible confidence levels $1-\delta$ is defined implicitly, whereas we provide an explicit interval, that depends only on $\varepsilon$ and $n$.
(5) \citet{monk} require $2n_\mathsf{O} \le K \le n/2$, meaning they allow at most $25\%$ of outliers, while we can handle up to $50\%$.
(6) They only prescribe a rough estimate of $K$, that might not be an integer.




\subsection{Concentration Bounds for MoU}

\begin{figure}[!t]
\begin{center}
\includegraphics[width=\columnwidth, page=3]{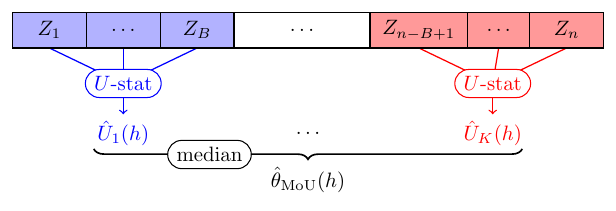}
\caption{The MoU estimator.}
\label{fig:mou}
\end{center}
\vspace{-0.3cm}
\end{figure}

Many machine learning problems can be formulated as the minimization of a certain $U$-statistic, an average over tuples of observations, generalizing the basic sample mean (one may refer to \citet{Lee90} for an account of the theory of $U$-statistics): ranking \citep{CLV08}, clustering, see \textit{e.g.} \citet{CLEM14}, or metric-learning \citep{VCB18} among others.
We recall that the $U$-statistic of degree $d\in\{1,\; \ldots,\; n\}$  with kernel $h:\mathbb{R}^d\rightarrow \mathbb{R}$, symmetric (\textit{i.e.} invariant under permutation of its arguments), square integrable w.r.t. $P^{\otimes d}$, denoting by $P$ the distribution of the random variable $Z$, and based on independent copies $Z_1,\; \ldots ,\; Z_n$ of $Z$  is given by:
\begin{equation}\label{eq:Ustat1}
\bar{U}_n(h)=\frac{1}{\binom{n}{d}}\sum_{1\leq i_1<\ldots<i_d\leq n}h(Z_{i_1},\; \ldots ,\; Z_{i_d}).
\end{equation}
As may be shown by a Lehmann-Scheff\'e argument, it is the unbiased estimator of the parameter $\theta(h)=\int h(z_1,\; \ldots,\; z_d)P(dz_1)\ldots P(dz_d)$ with minimal variance, given by (see \textit{e.g.} \citet{van2000asymptotic}):
\begin{equation*}
\frac{1}{\binom{n}{d}}\sum_{c=1}^d\binom{d}{c}\binom{n-d}{d-c}\zeta_c(h) \le \frac{d!}{n}\sum_{c=1}^d \binom{d}{c} \zeta_c(h),
\end{equation*}
where, for $1\leq c\leq d$, we have set $\zeta_c(h)=\mathrm{Var}(h_c(Z_1,\; \ldots,\; Z_c))$, with $h_c(z_1,\; \ldots,\; z_c)=\mathbb{E}[h(z_1,\; \ldots,\; z_c,\; Z_{c+1},\; \ldots,\; Z_d)]$ for  all $(z_1,\; \ldots,\; z_c)\in \mathbb{R}^c$.
%
%
As a single outlier affects $\binom{n-1}{d-1}$ terms among those averaged in \eqref{eq:Ustat1}, it is essential to design robust alternatives.
Medians-of-$U$-statistics (MoU) naturally extend the MoM approach by considering the median of $U$-statistics built on disjoint blocks $\mathcal{B}_1,\; \ldots,\; \mathcal{B}_K$ of size $B\geq d$ (see \citet{joly2016robust} for the case of degenerate $U$-statistics, or \citet{laforgue2019medians} for a general study on randomized, possibly overlapping, blocks).
The MoU estimator of $\theta(h)$ is defined as $\hat{\theta}_\mathrm{MoU}(h) = \mathrm{median}\big(\hat{U}_k(h),~k\le K\big)$, with
\begin{equation*}
\hat{U}_k(h)= \frac{1}{\binom{B}{d}} \sum_{i_1<\ldots<i_d \in \mathcal{B}_k} h(Z_{i_1}, \ldots, Z_{i_d}) \quad \text{for } k\le K.
\end{equation*}
See \Cref{fig:mou} for a depiction.
The next proposition details the concentration guarantees of $\hat{\theta}_\mathrm{MoU}(h)$ when the sample $\mathcal{S}_n$ it is based upon is contaminated according to \Cref{hyp:framework}.
The technical proof is detailed in \Cref{apx:proof_mou_1_sample}.

\begin{proposition}\label{prop:mou_1sample}
Suppose that sample $\mathcal{S}_n$ and mapping $\alpha$ satisfy \Cref{hyp:framework,hyp:config} respectively.
Define \mbox{functions} $\beta, \gamma, \Gamma, \Delta$ according to \Cref{hyp:config}, and set $\Sigma^2(h)$ as follows: $\Sigma^2(h) =  d! \sum_{c=1}^d \binom{d}{c}\zeta_c(h)$.
Then, for all $\delta \in [e^{-n/\beta(\varepsilon)}, e^{-n\alpha(\varepsilon)/\beta(\varepsilon)}]$, choosing $K = \lceil \beta(\varepsilon)\log(1/\delta)\rceil$, it holds with probability larger than $1-\delta$:
\begin{equation*}
\big|\hat{\theta}_{\mathrm{MoU}}(h) - \theta(h)\big| ~\le~ 4\sqrt{e}~\Sigma(h)~\gamma(\varepsilon)~\sqrt{\frac{1 + \log(1/\delta)}{n}}.
\end{equation*}

If in addition the essential supremum $\|h(Z_1, \ldots, Z_d)\|_\infty=\inf\{t\geq 0: \mathbb{P}\{\vert h(Z_1, \ldots, Z_d)\vert >t \}=0\}$ of the r.v. $|h(Z_1, \ldots, Z_d)|$ is finite and bounded by $M$, then for all $\delta \in ]0, e^{-4n\alpha(\varepsilon)}]$, choosing $K = \lceil \alpha(\varepsilon)n\rceil$, it holds with probability at least $1 - \delta$:
\begin{equation*}
\big|\hat{\theta}_{\mathrm{MoU}}(h) - \theta(h)\big| ~\le~ 4 \sqrt{d}~M~\Gamma(\varepsilon)~\sqrt{\frac{\log(1/\delta)}{n}}.
\end{equation*}

If furthermore $n_\mathsf{O}$ satisfies \Cref{ass:sub_linear}, the same $K$ gives:
\begin{align*}
\mathbb{E}\Big[ \big|\hat{\theta}_\mathrm{MoU}(h)& - \theta(h)\big|\Big]\\
&\le~ 2\sqrt{d}~M~\Gamma(\varepsilon)\left(4C_\mathsf{O}~\frac{\Delta(\varepsilon)}{n^{(1 - \alpha_\mathsf{O})/2}} + \sqrt{\frac{\pi}{n}}\right).
\end{align*}
\end{proposition}

\subsection{Concentration Bounds for Multisample MoU}

\begin{figure*}[!ht]
\vspace{-0.1cm}
\begin{center}
\begin{minipage}[r]{0.32\textwidth}
\includegraphics[width=\textwidth]{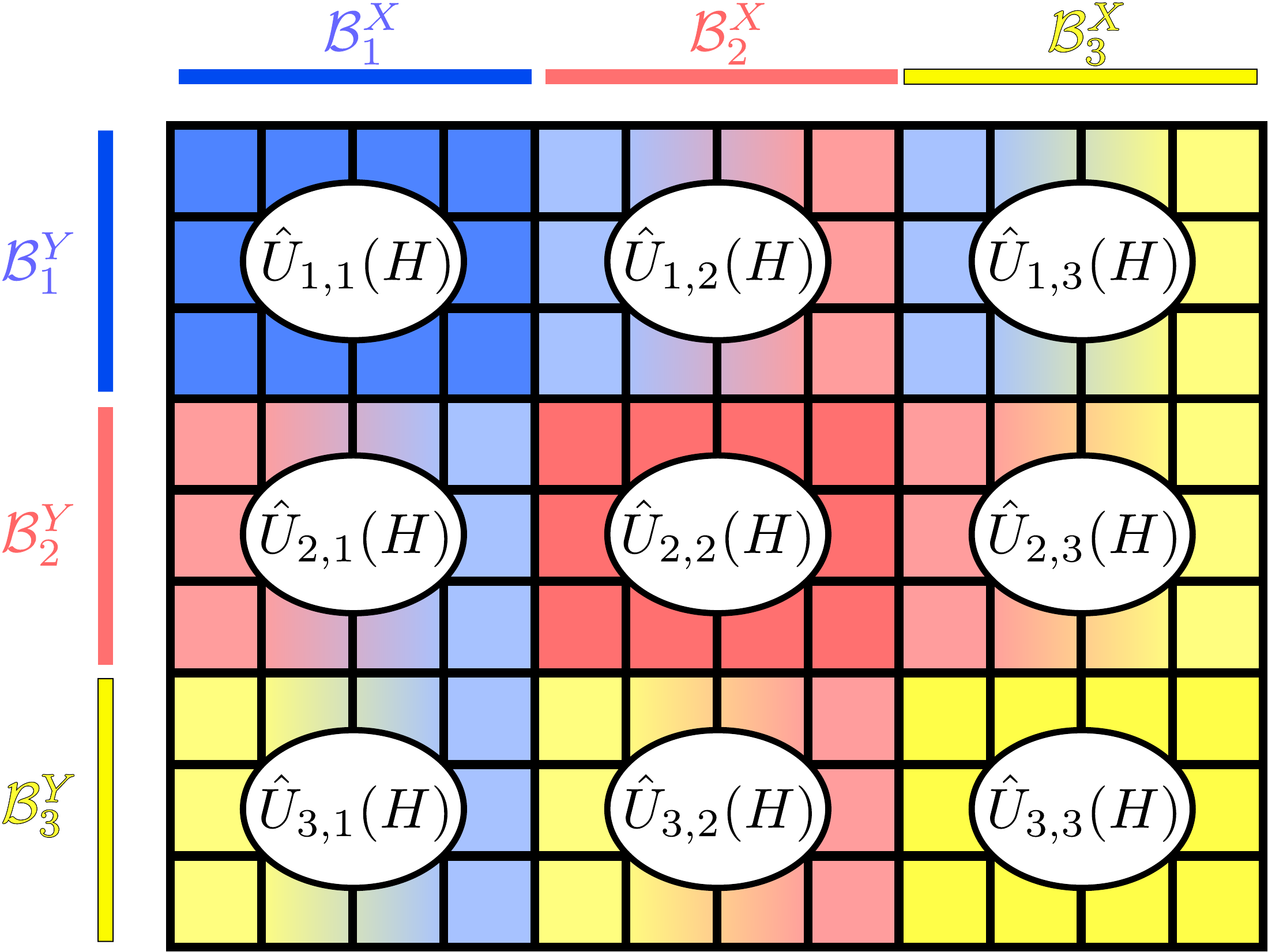}\vspace{0.35cm}
\caption{The $\mathrm{MoU}_2$ estimator.}
\label{fig:mou_2}
\end{minipage}
\hfill
\begin{minipage}[r]{0.32\textwidth}
\includegraphics[width=\textwidth]{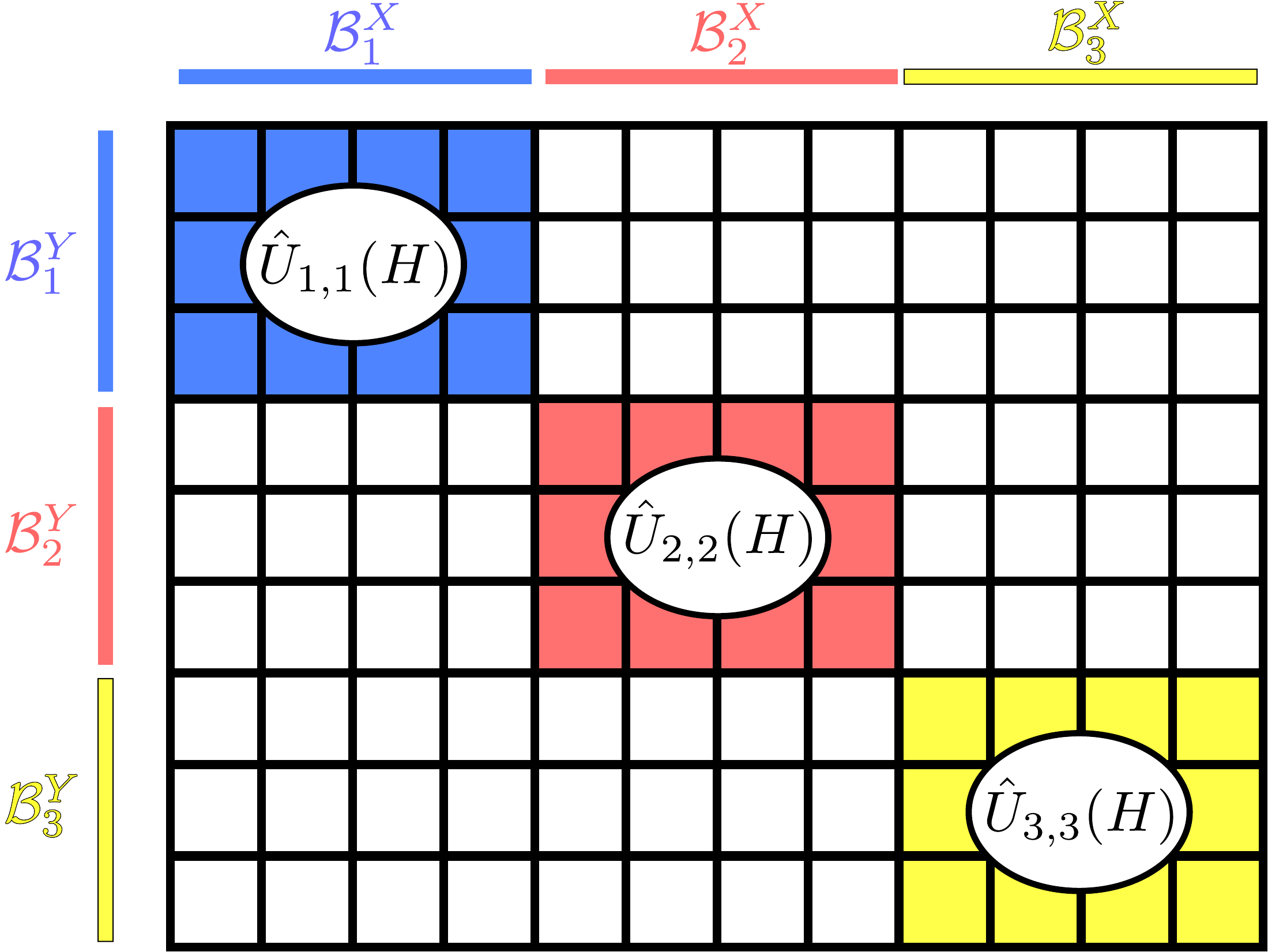}\vspace{0.35cm}
\caption{The $\mathrm{MoU}_2^\text{diag}$ estimator.}
\label{fig:mou_2_diag}
\end{minipage}
\hfill
\begin{minipage}[r]{0.3\textwidth}
\vspace{0.45cm}
\includegraphics[width=0.83\textwidth]{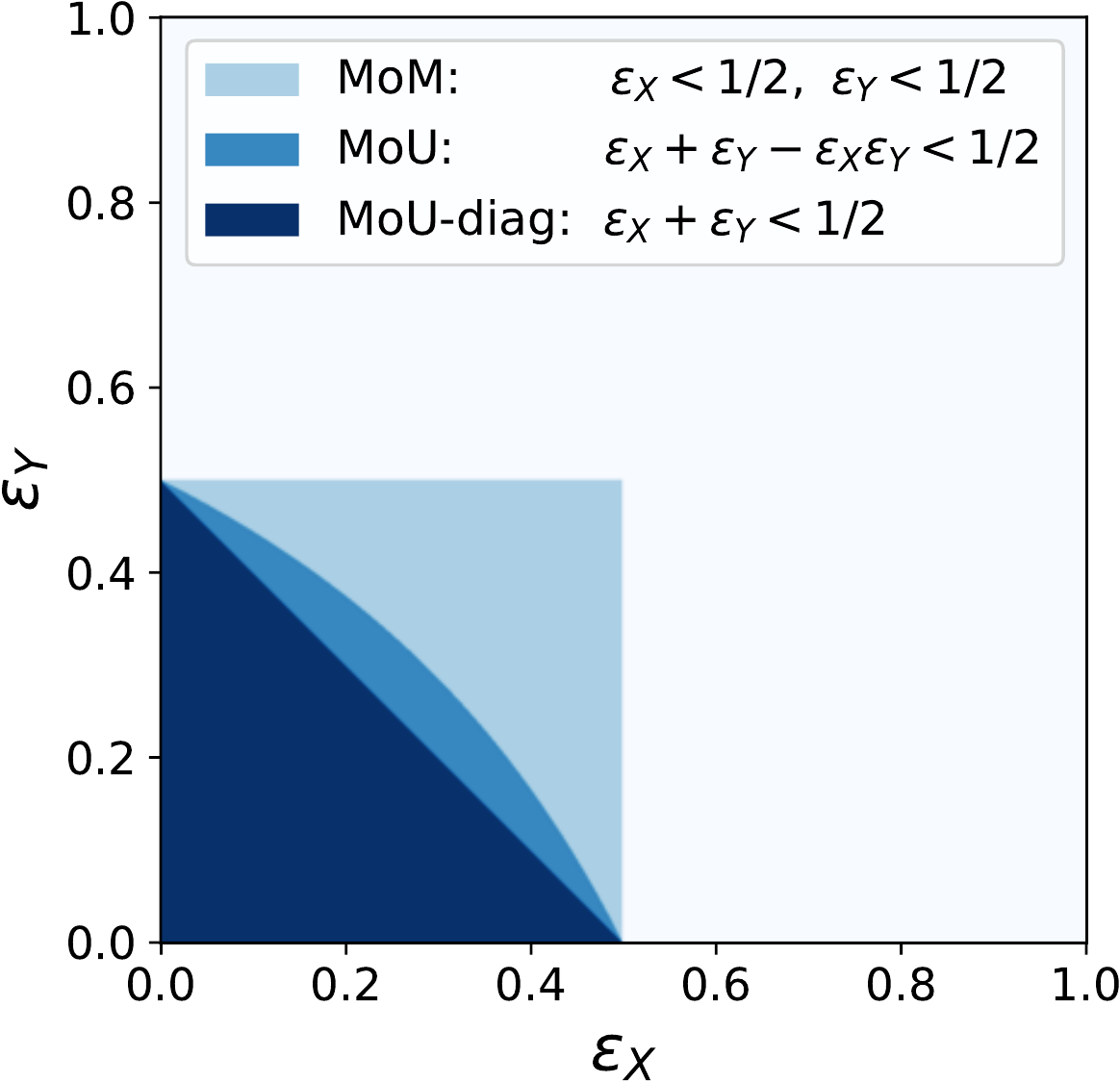}
\vspace{-0.1cm}
\caption{Outliers accepted.}
\label{fig:constraints}
\end{minipage}
\end{center}
\vspace{-0.4cm}
\end{figure*}

The notion of $U$-statistic can be readily extended to the multisample framework, see \citet{Lee90}.
For notational simplicity, we restrict ourselves to $2$-sample $U$-statistics of degrees $(1,1)$.
Extensions to $U$-statistics of arbitrary degrees and/or based on more than two samples are direct and detailed in \Cref{apx:ext_ustat}.
The $U$-statistic of degrees $(1,1)$ with kernel $H:\mathbb{R}^2\rightarrow \mathbb{R}$, square integrable w.r.t. $P \otimes Q$, denoting by $P$ and $Q$ the distributions of r.v. $X$ and $Y$ respectively, and based on two independent samples $\mathcal{S}_n^X =\{ X_1, \ldots , X_n\}$, and $\mathcal{S}_m^Y = \{Y_1, \ldots , Y_m\}$, composed respectively of $n\geq 1$ and $m\ge 1$ independent copies of $X$ and $Y$, is given by:
\begin{equation*}
\bar{U}_{n,m}(H)=\frac{1}{nm}\sum_{i=1}^n\sum_{j=1}^mH(X_{i},\; Y_{j}).
\end{equation*}
It is the unbiased estimator of~~$\theta(H)=\int\int H(x,y)P(dx)Q(dy)$ with minimal variance, given by:
\begin{align}\label{eq:var_ustat_multi}
\sigma^2_{n,m}(H) &=\frac{1}{nm}\sigma^2(H) + \frac{m-1}{nm} \sigma_1^2(H) + \frac{n-1}{nm} \sigma_2^2(H),\nonumber\\
&\le \frac{\sigma^2(H) + \sigma_1^2(H) + \sigma_2^2(H)}{n \wedge m},
\end{align}
where $\sigma^2(H) = \mathrm{Var}(H(X,Y))$, $\sigma_1^2(H) = \mathrm{Var}(H_1(X))$ and $\sigma_2^2(H) = \mathrm{Var}(H_2(Y))$, with $H_1(x) = \mathbb{E}\left[H(x, Y)\right]$ and $H_2(y) = \mathbb{E}\left[H(X, y)\right]$.
Similarly to MoM, each sample is divided into $K_X$ (respectively $K_Y$) disjoint blocks of size $B_X = \lfloor n/K_X\rfloor$ (respectively $B_Y = \lfloor m/K_Y\rfloor$).
The Median-of-(two-sample)-$U$-statistics estimator is then given by $\hat{\theta}_{\mathrm{MoU}_2}(H) = \mathrm{median}\big(\hat{U}_{k, l}(H),~k, l \le K_X, K_Y\big)$, with
\begin{equation*}
\hat{U}_{k, l}(H) = \hspace{-0.3cm}\sum_{i, j \in \mathcal{B}^X_k \times \mathcal{B}^Y_l} \hspace{-0.2cm} \frac{H(X_i, X_j)}{B_X B_Y}, \quad \text{for } k, l \le K_X, K_Y.
\end{equation*}

Refer to \Cref{fig:mou_2} for a visual interpretation in the particular case $K_X = K_Y = 3$.
For $\mathrm{MoU}_2$, the total number of blocks created is thus $K_X K_Y$, while the number of corrupted ones is always lower than $n_\mathsf{O} K_Y + m_\mathsf{O} K_X - n_\mathsf{O} m_\mathsf{O}$.
As we still want at least twice more blocks than possibly corrupted ones, the constraint on $K_X$ and $K_Y$ can be expressed as:
\begin{equation*}
2(\varepsilon_X + \varepsilon_Y - \varepsilon_X\varepsilon_Y)nm < K_X K_Y \le nm.
\end{equation*}

The proportions of outliers $\varepsilon_X$ and $\varepsilon_Y$ for which we are able to derive statistical guarantees should therefore satisfy $\varepsilon_X + \varepsilon_Y - \varepsilon_X\varepsilon_Y < 1/2$.
This is a stronger requirement than for MoM, see \Cref{fig:constraints}.
The next proposition then details the concentration properties of $\mathrm{MoU}_2$ under this assumption.

\begin{proposition}\label{prop:mou_2sample}
Suppose that both samples $\mathcal{S}_n^X$ and $\mathcal{S}_m^Y$ and mapping $\alpha$ satisfy \Cref{hyp:framework,hyp:config} respectively.
Define functions $\beta, \gamma, \Gamma, \Delta$ according to \Cref{hyp:config}.
Let $\varepsilon_X$ and $\varepsilon_Y$ be such that $\tilde{\varepsilon} \coloneqq \varepsilon_X + \varepsilon_Y -\varepsilon_X\varepsilon_Y$ is strictly smaller than $1/2$.
Then, for all $\delta \in [2\max(e^{-n\beta_X}, e^{-m\beta_Y})$, $2\min(e^{-n\sqrt{\alpha}(\tilde{\varepsilon})/\beta_X}, e^{-m\sqrt{\alpha}(\tilde{\varepsilon})/\beta_Y})]$, choosing $K_X = \left\lceil \beta_X \log(2/\delta)\right\rceil$, and $K_Y = \left\lceil \beta_Y \log(2/\delta)\right\rceil$, it holds with probability at least $1-\delta$:
\begin{equation*}
\big|\hat{\theta}_{\mathrm{MoU}_2}(H) - \theta(H)\big| ~\le~ 12\sqrt{3}~\Sigma(H)\gamma(\tilde{\varepsilon})\sqrt{\frac{1 + \log(2/\delta)}{n \wedge m}},
\end{equation*}
with the notation $\Sigma^2(H) = \sigma^2(H) + \sigma_1^2(H) + \sigma_2^2(H)$, $\beta_{Z} = \frac{18~\eta^2(\tilde{\varepsilon})}{\eta_{Z}(2\eta(\tilde{\varepsilon}) - 1)^2}$, and $\eta_{Z} = 1 - \frac{\varepsilon_{Z}}{\sqrt{\alpha}(\tilde{\varepsilon})}$, for $Z=X,\; Y$.
\end{proposition}



The technical proof is detailed in \Cref{apx:cross_block}, and is made significantly more involved due to the introduction of dependent random variables, see the $\hat{U}_{k,l}(H)$ in \Cref{fig:mou_2}.
The conditional Hoeffding's inequality then provides an alternative to the Binomial concentration, with the major drawback that it does not allow for a sharp analysis if one further assumes that $\|H(X, Y)\|_\infty$ is finite, see also the discussion in \Cref{rmk:hoeffding}.
As a result, \Cref{prop:mou_2sample} must be restricted to guarantees on the restricted range of confidence levels.
Notice that randomized extensions considered in \citet{laforgue2019medians} rely on Hoeffding's inequality as well, and consequently suffer from the same restriction.
To overcome this limitation, an alternative consists in removing the dependence between the $U$-statistics, at the cost of a loss of information though.

Indeed, getting independent $U$-statistics might be easily achieved, by considering only the diagonal blocks as in \Cref{fig:mou_2_diag}.
This procedure however results in an important loss of information, since a large portion of the grid remains unexplored.
Another drawback of this approach is that it forces to set $K_X = K_Y = K$.
Overall, this estimator, denoted $\hat{\theta}_{\mathrm{MoU}_2^\text{diag}}(H)$ is given by
\begin{equation}\label{eq:mou_diag}
\hat{\theta}_{\mathrm{MoU}_2^\text{diag}}(H) = \mathrm{median}\left(\hat{U}_{k, k}(H),~~k \le K\right).
\end{equation}
The constraint on $K$ then becomes: $2(n_\mathsf{O} + m_\mathsf{O}) < K \le \min(n, m)$.
Obviously, as soon as $m \le 2n_\mathsf{O}$ this cannot be satisfied.
To avoid such problems, we shall assume that $n = m$, see the discussion at the end of the section.
We now analyze the concentration properties of estimator \eqref{eq:mou_diag}.

\begin{proposition}\label{prop:mou_diag}
Suppose that samples $\mathcal{S}_n^X$ and $\mathcal{S}_m^Y$ and mapping $\alpha$ satisfy \Cref{hyp:framework,hyp:config} respectively.
Define functions $\beta, \gamma, \Gamma, \Delta$ according to \Cref{hyp:config}, and assume that $\varepsilon_X + \varepsilon_Y < 1/2$.
Then, for all $\delta \in [e^{-n/\beta(\varepsilon_X + \varepsilon_Y)}, e^{-n\alpha(\varepsilon_X + \varepsilon_Y)/\beta(\varepsilon_X + \varepsilon_Y)}]$, with $K = \left\lceil \beta(\varepsilon_X + \varepsilon_Y)\log(1/\delta)\right\rceil$, it holds w.p.a.l. $1-\delta$:
\begin{align*}
\big|\hat{\theta}_{\mathrm{MoU}_2^\text{diag}}(H) &- \theta(H)\big|\\
&\le~ 4\sqrt{e}~\Sigma(H)~\gamma(\varepsilon_X + \varepsilon_Y)~\sqrt{\frac{1 + \log(1/\delta)}{n}}.
\end{align*}

If in addition $\|H(X, Y)\|_\infty$ is finite and upper bounded by $M$, then for  all $\delta \in ]0, e^{-4n\alpha(\varepsilon_X + \varepsilon_Y)}]$, choosing $K = \lceil \alpha(\varepsilon_X + \varepsilon_Y)n\rceil$, it holds with probability at least $1 - \delta$:
\begin{align*}
\big|\hat{\theta}_{\mathrm{MoU}_2^\text{diag}}(H) - \theta(H)\big| \le~ 8M~\Gamma(\varepsilon_X + \varepsilon_Y)~\sqrt{\frac{\log(1/\delta)}{n}}.
\end{align*}

If furthermore $n_\mathsf{O}$ and $m_\mathsf{O}$ satisfy \Cref{ass:sub_linear}, the same $K$ gives:
\begin{align*}
\mathbb{E}&\Big[ \big|\hat{\theta}_{\mathrm{MoU}_2^\text{diag}}(H) - \theta(H)\big|\Big]\\
&\le~ 4M~\Gamma(\varepsilon_X + \varepsilon_Y)\left(4\sqrt{2}~C_\mathsf{O}~\frac{\Delta(\varepsilon_X + \varepsilon_Y)}{n^{(1 - \alpha_\mathsf{O})/2}} + \sqrt{\frac{\pi}{n}}\right).
\end{align*}
\end{proposition}

The proof can be found in \Cref{apx:mou_diag}.
Notice that the constraint $n = m$ can be relaxed, as long as $2(n_\mathsf{O} + m_\mathsf{O}) \le \min(n, m)$ still holds.
However, the case $n=m$ is the only one documented in MoM's literature to our knowledge \citep{monk}, while it nicely exhibits the critical point $\varepsilon_X + \varepsilon_Y = 1/2$.
When estimating Integral Probability Metrics \citep{Bharath2012}, one typically relies on two-sample $U$-statistics, built upon kernels of the form $H_\phi(X, Y) = \phi(X) - \phi(Y)$, for $\phi$ in the functional set considered.
Hence, one might use a MoM-MoM estimate, instead of a MoU$_{2}$ or a MoU$_2^\text{diag}$ estimate (see \citet{staerman2020ot} for an application to the estimation of the \mbox{1-Wasserstein} distance).
The corresponding proportions of outliers admitted would be $\varepsilon_X < 1/2$, and $\varepsilon_Y < 1/2$, that represents a less stringent constraint, as shown in \Cref{fig:constraints}.
For $p$-sample $U$-statistics this constraints would write as $\|\bm{\varepsilon}\|_\infty < 1/2$ for a MoM-based estimate, and $\|\bm{\varepsilon}\|_1 < 1/2$ for MoU$_p$, with $\bm{\varepsilon} = (\varepsilon_1, \ldots, \varepsilon_p)$ the vector containing the $p$ samples proportions of outliers.

\section{Statistical Guarantees for Pairwise Learning in the Presence of Outliers}\label{sec:learning}

A simple and meaningful way to illustrate the relevance of MoM-based estimators in the presence of outliers is to use them for revisiting the Empirical Risk Minimization paradigm (ERM, see \textit{e.g.} \citet{DGL96}).
Consider a generic supervised learning problem, defined by a pair of input/output random variables $Z = (X, Y) \in \mathcal{Z} = \mathcal{X} \times \mathcal{Y}$ with unknown distribution $P$, a hypothesis set $\mathcal{G} \subset \mathcal{Y}^\mathcal{X}$, and a loss function $\ell\colon\mathcal{G}\times\mathcal{Z}\rightarrow \mathbb{R}_+$.
ERM then consists in substituting the unknown risk $\mathbb{E}_P\left[\ell(g, Z)\right]$ by its empirical version based on sample $\mathcal{S}_n$, and solving the optimization problem $\min_{g \in \mathcal{G}}(1/n) \sum_{i=1}^n \ell(g, Z_i)$.
When $\mathcal{S}_n$ is possibly contaminated, a natural idea to robustify ERM is to solve instead $\min_{g \in \mathcal{G}} \mathrm{MoM}_{\mathcal{S}_n}[\ell(g, Z)]$.
This approach, explored in \citet{lecue2018robust} for standard MoMs by means of \textit{ad hoc} Rademacher complexities tailored to outliers, is referred to as MoM-minimization.
This section builds upon the concentration bounds established in \Cref{sec:revisit} to extend these ideas to pairwise learning problems, with a simpler formalism based on the Vapnik-Chervonenkis dimension.
Consider now a hypothesis set $\mathcal{G} \subset \{-1, +1\}^{\mathcal{X} \times \mathcal{X}}$, and a symmetric loss function $\ell\colon \mathcal{G}\times\mathcal{Z}^2 \rightarrow \mathbb{R}_+$.
Let $Z'$ denote an independent copy of $Z$, and set $\ell_g(Z, Z') = \ell(g, Z, Z')$.
Our goal is to find a decision rule $g^*$ that minimizes over~$\mathcal{G}$ $\mathcal{R}(g) = \mathbb{E}_{Z, Z'}\left[\ell_g(Z, Z')\right]$ .
A classical example covered by this setting is \emph{ranking}, where one is typically interested in predicting if some object $X$ is preferred over some other object $X'$.
We study the performance of the MoU-minimizer $\hat{g}_{\mathrm{MoU}} = \argmin_{g \in \mathcal{G}}~\mathrm{MoU}_{\mathcal{S}_n}(\ell_g)$, where
\begin{align*}
\mathrm{MoU}_{\mathcal{S}_n}\left(\ell_g\right) = \text{median}\Big(\sum_{i<j \in \mathcal{B}_1}& \ell_g(Z_i, Z_j),\\
&\ldots \sum_{i<j \in \mathcal{B}_K} \ell_g(Z_i, Z_j)\Big).
\end{align*}
The following two assumptions on the hypothesis set and the loss are required to our analysis.

\begin{assumption}\label{hyp:vc_dim}
The hypothesis space $\mathcal{G}$ considered has finite VC dimension $\textsc{VC}_\text{dim}(\mathcal{G})$.
\end{assumption}

\begin{assumption}\label{hyp:bounded_loss}
There exists $M > 0$ such that it holds $\ell(g, Z, Z') \le M$ almost surely.
\end{assumption}

\Cref{hyp:vc_dim,hyp:bounded_loss} are standard in statistical learning.
One typically has $M=1$ for the $0\text{-}1$ loss $\ell\colon (g, Z, Z') \mapsto \mathbbm{1}\{(g(X, X')(Y - Y') \le 0\}$.
Notice that if $\mathcal{Y}$ is bounded, any convex relaxation of the latter also fits.
We again stress that \Cref{hyp:bounded_loss} only applies to the inliers, \textit{i.e.} to the realizations of $Z$ and $Z'$, not necessarily to the outliers.
The next theorem characterizes $\hat{g}_\mathrm{MoU}$'s generalization capacity.

\begin{theorem}\label{thm:mou_min}
Suppose that sample $\mathcal{S}_n$ and mapping $\alpha$ satisfy \Cref{hyp:framework,hyp:config} respectively.
Define functions $\Gamma, \Delta$ according to \Cref{hyp:config}.
Assume furthermore that $\mathcal{G}$ and $\ell$ satisfy \Cref{hyp:vc_dim,hyp:bounded_loss} respectively.
Then, for all $\delta \in [0, e^{-4\Delta^2(\varepsilon)n_\mathsf{O}}]$, choosing $K = \lceil \alpha(\varepsilon)n\rceil$, it holds with probability at least $1-\delta$:
\begin{align*}
\mathcal{R}(&\hat{g}_\mathrm{MoU}) - \mathcal{R}(g^*)\\
&\le 8\sqrt{2}M~\Gamma(\tau)\sqrt{\frac{\textsc{VC}_\text{dim}(\mathcal{G})(1 + \log(n)) +\log(1/\delta)}{n}}.
\end{align*}
\end{theorem}

\Cref{thm:mou_min} is proved by combining the second claim of \Cref{prop:mou_1sample} with the complexity assumption on $\mathcal{G}$, details can be found in \Cref{apx:rade}.
We emphasize on the generic nature of the bounds established in \Cref{sec:revisit}.
This key property allows to efficiently combine them with various complexity assumptions on $\mathcal{G}$.
A second generalization bound based upon an entropic control of $\mathcal{G}$ is for instance proposed in \Cref{apx:chaining}.
In contrast, the guarantees in \citet{lecue2018robust} uses an \textit{ad hoc} Rademacher complexity  specifically tailored to their needs.
If VC dimensions are also used in \citet{depersin2020robust}, we emphasize that it is for estimation purposes, that do not relate to the learning bounds established in \Cref{thm:mou_min}.


From an algorithmic point of view, computing decision functions with guarantees similar to that in \Cref{thm:mou_min} can be done through MoU Gradient Descent (MoU-GD).
It is an pairwise adaptation of the algorithm proposed in \citet{lecue2018robust}, that can be described as follows.
For simplicity, we assume that $\mathcal{G}$ is a parametric hypothesis set of dimension $p$, \textit{i.e.} for every $g\in \mathcal{G}$ there exists $u \in \mathbb{R}^p$ such that $g = g_u$.
MoU-GD then revisits minibatch Gradient Descent in the following way.
At each step, the dataset is partitioned, and (pairwise) risk estimates are computed on each block.
The block with the median risk is selected, and a minibatch Gradient Descent step is computed, with the median block acting as the minibatch.
This is repeated until convergence.
The approach is formally detailed in \Cref{alg:gd}.
Observe that the partition needs to be randomized at each iteration in order to avoid local minima, see Remark 5 in \citet{lecue2018robust}.
Under standard convexity assumptions, we now show that the output of \Cref{alg:gd} converges towards $\hat{g}_\mathrm{alg}$, that enjoys the guarantees established in \Cref{thm:mou_min}.

\begin{theorem}\label{thm:alg}
Suppose that the assumptions of \Cref{thm:mou_min} hold, and that pairwise adaptations of the assumptions of Theorem 3 in \citet{lecue2018robust} hold.
Then, the output of \Cref{alg:gd} converges almost surely towards $\hat{g}_\mathrm{alg}$, that satisfies with probability at least $1 - \delta$:
\begin{align*}
\mathcal{R}(&\hat{g}_\mathrm{alg}) - \mathcal{R}(g^*)\\
&\le 8\sqrt{2}M~\Gamma(\tau)\sqrt{\frac{\textsc{VC}_\text{dim}(\mathcal{G})(1 + \log(n)) +\log(1/\delta)}{n}}.
\end{align*}
\end{theorem}

Due to space limitation, the explicit assumptions are detailed in \Cref{apx:thm_alg}, along with the proof of \Cref{thm:alg}.

\begin{figure}[!t]
\begin{center}
\includegraphics[width=0.86\columnwidth]{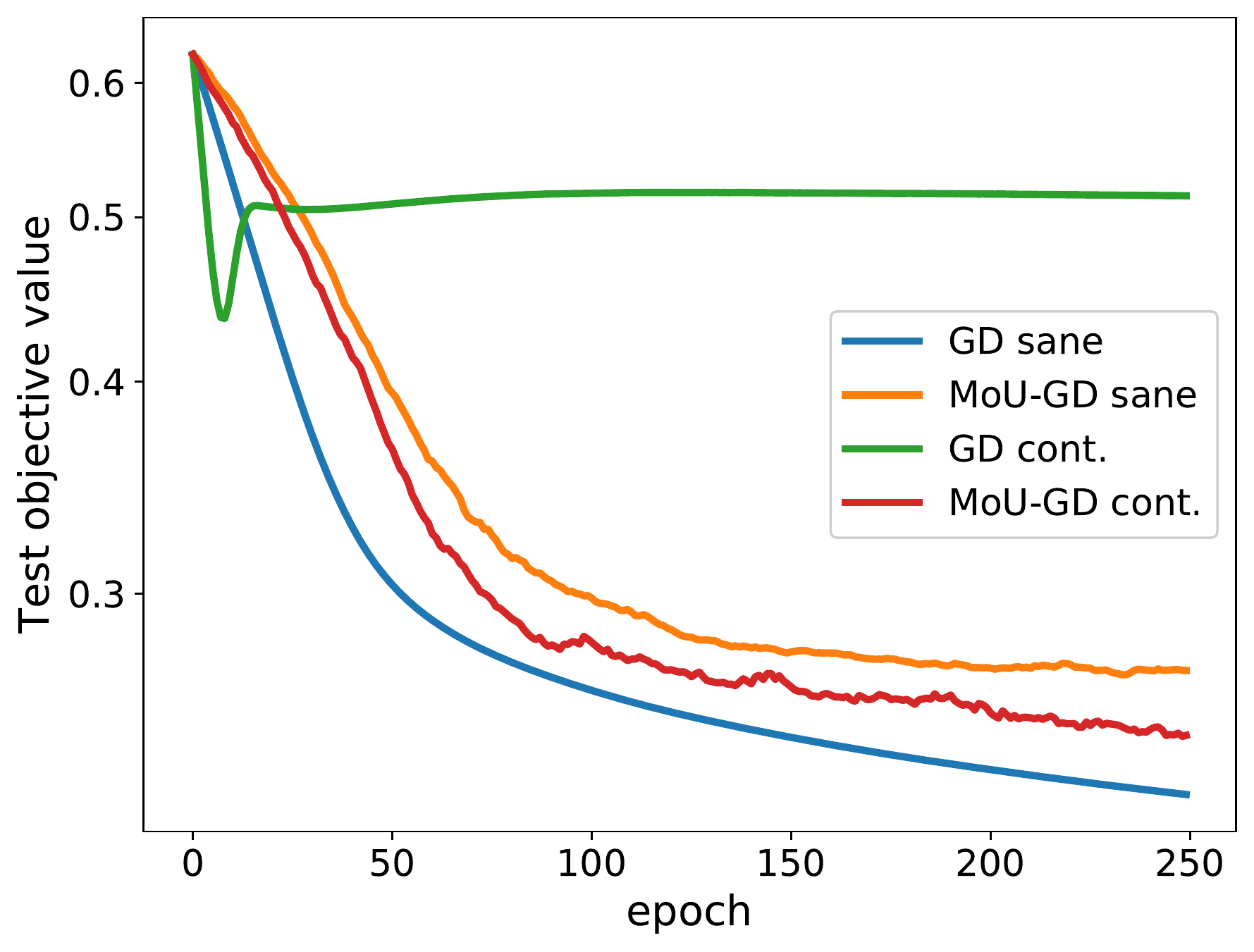}
\caption{The MoU Gradient Descent.}
\label{fig:mou_gd}
\end{center}
\vspace{-0.3cm}
\end{figure}

Empirically, \Cref{alg:gd} behaves in accordance with the theory.
\Cref{fig:mou_gd} shows the test trajectories of standard and MoU-GDs learned on sane and contaminated datasets: the contaminated GD converges towards a poor minimizer (w.r.t. the sane test data), while the MoU-GDs are insensitive to contamination and exhibit performances close to that of the sane GD.
More experiments can be found in \Cref{apx:expes}.

\begin{algorithm}[!t]
\SetKwInOut{Input}{input}
\caption{MoU Gradient Descent~~(MoU-GD)}
\Input{~~$\mathcal{S}_n$,~~$K$,~~$T \in \mathbb{N}^*$,~~$(\gamma_t)_{t \le T} \in \mathbb{R}_+^T$,~~$u_0 \in \mathbb{R}^p$}
\vspace{0.1cm}
\For{epoch from $1$ to $T$}{\vspace{0.1cm}

    \tcp{Randomly partition the data}
    Choose a random permutation $\pi$ of $\{1, \ldots, n \}$\vspace{0.15cm}

    Build a partition $B_1, \ldots, B_k$ of $\{\pi(1), \ldots, \pi(n)\}$

    \vspace{0.1cm}
    \tcp{Select block with median risk}
    \For{$k \le K$}{

        $\hat{U}_{B_k} = \sum_{i < j \in B_k^2}~\ell(g_{u_t}, Z_i, Z_j)$
        }

    Set $B_\text{med}$ s.t. $\hat{U}_{B_\text{med}} = \text{median}(\hat{U}_{B_k}, \ldots \hat{U}_{B_K})$

    \vspace{0.1cm}
    \tcp{Gradient step}
    $u_{t+1} = u_t - \gamma_t \sum_{i < j \in B_k^2} \nabla_{u_t}\ell(g_{u_t}, Z_i, Z_j)$
    }

\vspace{0.1cm}
\Return{$u_T$}
\label{alg:gd}
\end{algorithm}

\section{Conclusion}

Widely analyzed and proved valid in the context of heavy-tailed data, the Median-of-Means (MoM) estimator is now the subject of analyses under the Huber's contamination model.
%
%
The present article offers an exhaustive view of its robustness properties under this regime, and proposes several concentration bounds with clear dependence on the proportions of outliers $\varepsilon$ and the number of blocks $K$, that can be extended to (multisample) $U$-statistics.
%
These bounds are incidentally shown to supply a sound theoretical basis for the reliability of MoM-based learning techniques when the training data are possibly contaminated in part by outliers with arbitrary distribution.

\clearpage
\bibliographystyle{apalike}
\bibliography{ref}
\clearpage

\appendix
\onecolumn

\section{Summary: the different estimators considered in the present article}

\begin{figure*}[!ht]
\begin{center}
\begin{subfigure}[b]{0.45\textwidth}
\includegraphics[width=\textwidth, page=1]{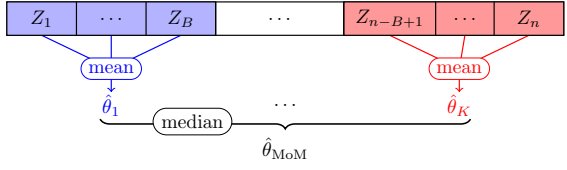}
\caption{The MoM Estimator.}
\end{subfigure}
\hfill
\begin{subfigure}[b]{0.45\textwidth}
\includegraphics[width=\textwidth, page=3]{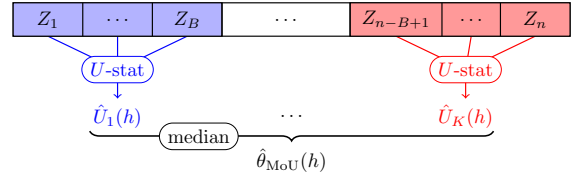}
\caption{The MoU Estimator.}
\end{subfigure}
\\[0.5cm]
\begin{subfigure}[b]{0.45\textwidth}
\includegraphics[width=\textwidth]{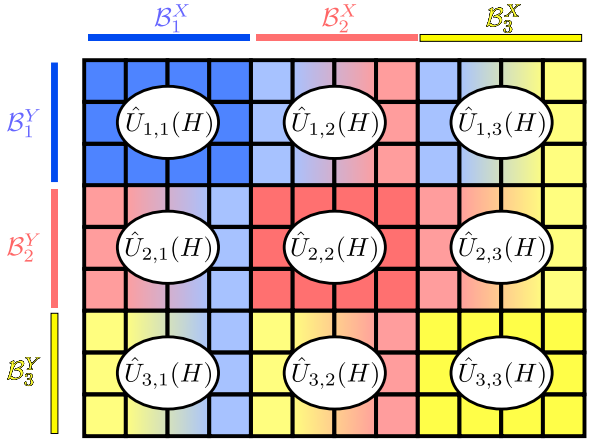}
\caption{The MoU$_2$ Estimator.}
\end{subfigure}
\hfill
\begin{subfigure}[b]{0.45\textwidth}
\includegraphics[width=\textwidth]{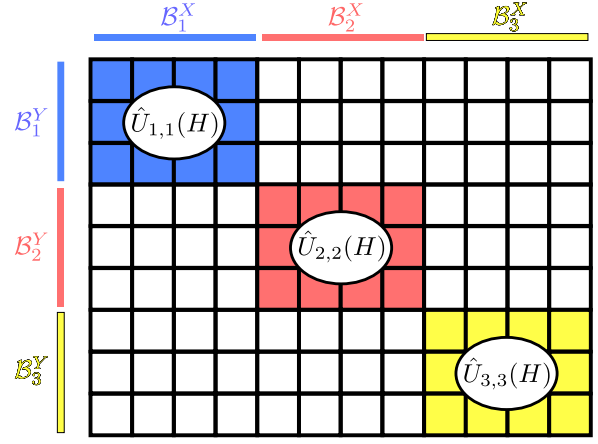}
\caption{The MoU$_2^\text{diag}$ Estimator.}
\end{subfigure}
\caption{The estimators considered in the article.}
\end{center}
\vspace{-0.3cm}
\end{figure*}


\section{Additional Tables and Figures}
\label{apx:plots}

\begin{table}[!ht]
\begin{center}
\begin{tabular}{c|ccccccc}\toprule
& $\alpha(\varepsilon)$ &  $\beta(\varepsilon)$ & $\gamma(\varepsilon)$ & $\Gamma(\varepsilon) $ & $\Delta(\varepsilon)$ & $\eta(\varepsilon)$\\\midrule\xrowht{35pt}
& $\alpha(\varepsilon)$ & $\displaystyle \frac{2\alpha(\varepsilon)}{\alpha(\varepsilon) - 2\varepsilon}$ &  $\displaystyle \frac{\sqrt{\alpha(\varepsilon)}(\alpha(\varepsilon) - \varepsilon)}{(\alpha(\varepsilon) - 2\varepsilon)^{3/2}}$ & $\displaystyle\sqrt{\frac{\alpha(\varepsilon)}{\alpha(\varepsilon) - 2\varepsilon}}$ & $\displaystyle\sqrt{\frac{\alpha(\varepsilon)}{\varepsilon}}$ & $\displaystyle\frac{\alpha(\varepsilon) - \varepsilon}{\alpha(\varepsilon)}$\\\hline\xrowht{35pt}
{\sc Arithmetic} & $\displaystyle\frac{1 + 2\varepsilon}{2}$ &  $\displaystyle\frac{2(1 + 2\varepsilon)}{1-2\varepsilon}$ & $\displaystyle \frac{\sqrt{1+2\varepsilon}}{(1 - 2\varepsilon)^{3/2}}$ & $\displaystyle\frac{\sqrt{1+2\varepsilon}}{\sqrt{1 - 2\varepsilon}}$ & $\displaystyle\sqrt{\frac{1+2\varepsilon}{2\varepsilon}}$ & $\displaystyle\frac{1}{1 + 2\varepsilon}$\\\hline\xrowht{35pt}
{\sc Geometric} & $\displaystyle\sqrt{2\varepsilon}$ &  $\displaystyle \frac{2(1 + \sqrt{2\varepsilon})}{1 - 2\varepsilon}$ & $\displaystyle\frac{(2 - \sqrt{2\varepsilon})(1 + \sqrt{2\varepsilon})^{3/2}}{2(1 - 2\varepsilon)^{3/2}}$ & $\displaystyle\frac{\sqrt{1+\sqrt{2\varepsilon}}}{\sqrt{1 - 2\varepsilon}}$ & $\sqrt[4]{2/\varepsilon}$ & $\displaystyle\frac{2 - \sqrt{2\varepsilon}}{2}$\\\hline\xrowht{35pt}
{\sc Harmonic} & $\displaystyle\frac{4\varepsilon}{1 + 2\varepsilon}$ & $\displaystyle\frac{4}{1 - 2\varepsilon}$ & $\displaystyle\frac{3 - 2\varepsilon}{\sqrt{2}(1 - 2\varepsilon)^{3/2}}$ & $\displaystyle\frac{\sqrt{2}}{\sqrt{1 - 2\varepsilon}}$ & $\displaystyle\sqrt{\frac{4}{1+2\varepsilon}}$ & $\displaystyle\frac{3 - 2\varepsilon}{4}$\\\hline\xrowht{35pt}
{\sc Polynomial} & $\displaystyle\varepsilon\Big(\frac{5}{2} - \varepsilon\Big)$ &  $\displaystyle\frac{2(5 - 2\varepsilon)}{1 - 2\varepsilon}$ & $\displaystyle\frac{(3 - 2\varepsilon)\sqrt{5 - 2\varepsilon}}{(1 - 2\varepsilon)^{3/2}}$ & $\displaystyle\frac{\sqrt{5 - 2\varepsilon}}{\sqrt{1 - 2\varepsilon}}$ & $\displaystyle\sqrt{\frac{5 - 2\varepsilon}{2}}$ & $\displaystyle\frac{3 - 2\varepsilon}{5 - 2\varepsilon}$\\\bottomrule
\end{tabular}
\caption{Different upper bounds $\alpha$ and corresponding functions $\beta, \gamma, \Gamma, \Delta, \eta$.}
\label{tab:summary}
\end{center}
\end{table}
\vspace{1cm}


%








\begin{figure*}[!ht]
\vspace{1cm}
\begin{center}
\includegraphics[width=0.48\textwidth]{Figures/plot_alpha}\hfill
\includegraphics[width=0.48\textwidth]{Figures/plot_gamma}\\
\includegraphics[width=0.48\textwidth]{Figures/plot_range}\hfill
\includegraphics[width=0.48\textwidth]{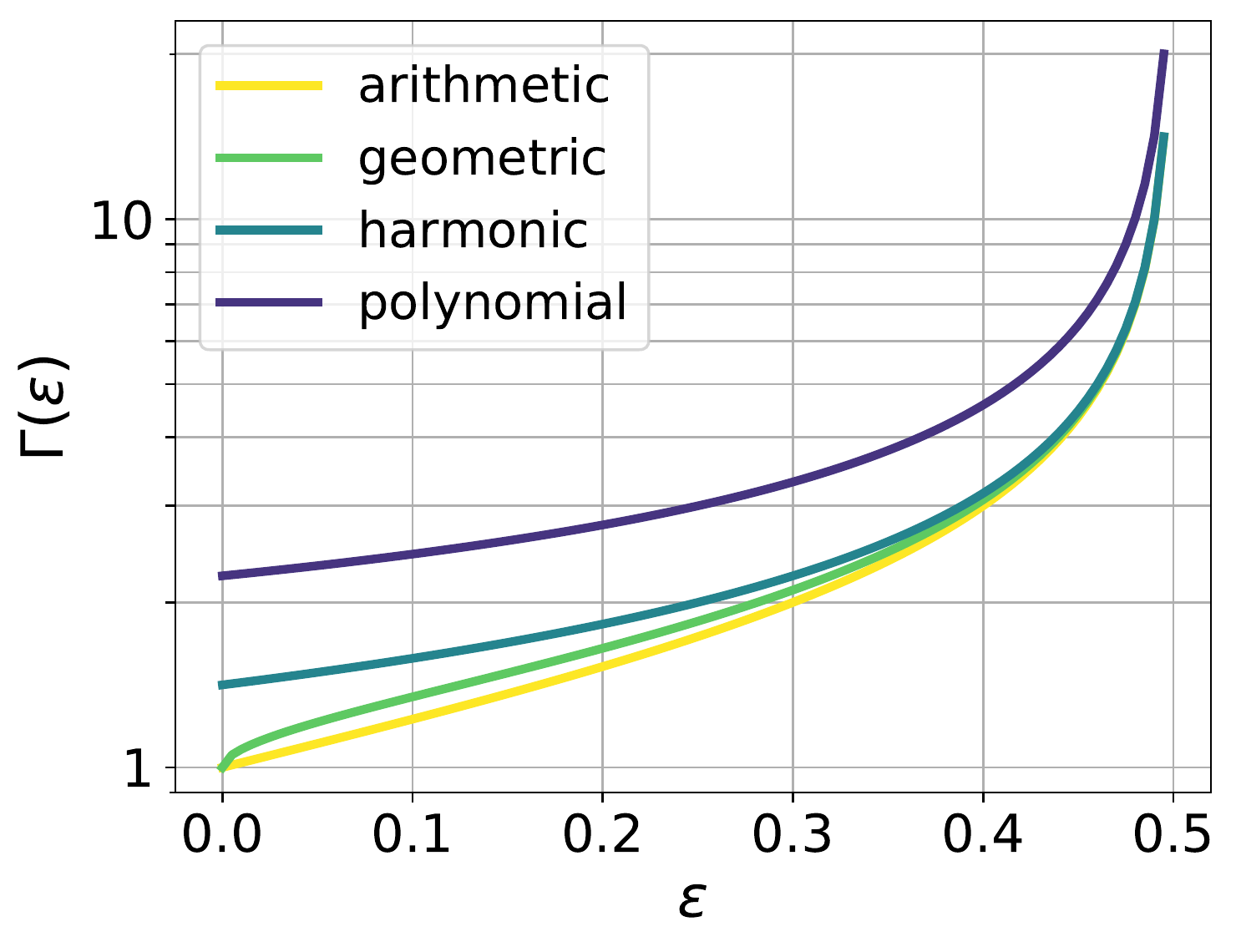}\\
\includegraphics[width=0.48\textwidth]{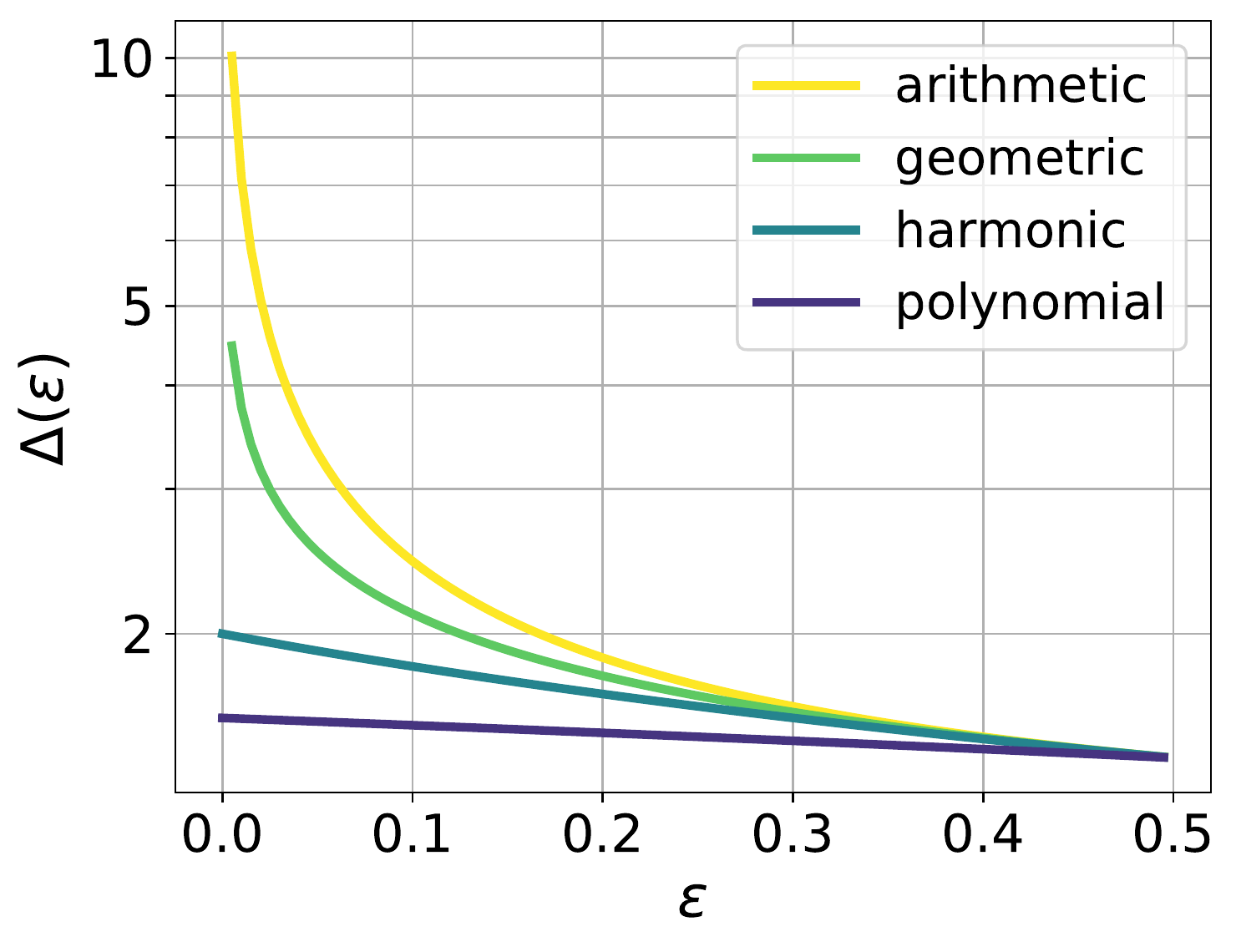}\hfill
\includegraphics[width=0.48\textwidth]{Figures/plot_Gamma_Delta}
\caption{Influence of the chosen mapping $\alpha$ on the constants.}
\label{fig:influence_apx}
\end{center}
\vspace{-0.3cm}
\end{figure*}

\clearpage


\section{Technical Proofs}
\label{apx:proof}

In this section are detailed the proofs of the theoretical claims stated in the core article.

\subsection{Proof of \Cref{prop:mom}}
\label{apx:proof_mom}

Roughly speaking, the median has the same behavior as that of a majority of observations.
Similarly, the MoM has the same behavior as that of a majority of blocks.
In presence of outliers, the key point consists in focusing on \emph{sane} blocks only, \textit{i.e.} on blocks that do not contain a single outlier, since no prediction can be made about blocks \emph{hit} by an outlier, in absence of any structural assumption concerning the contamination.
One simple way to ensure the sane blocks to be in (almost) majority is to consider twice more blocks than outliers.
Indeed, in the worst case scenario each outlier contaminates one block, but the sane ones remain more numerous.
Let $K$ denote the total number of blocks chosen, $K_\mathsf{O}$ the number of blocks containing at least one outlier, and $K_\mathsf{S}$ the number of sane blocks containing no outlier.
The crux of our proofs then consists in determining some $\eta > 1/2$ (that eventually depends on $\varepsilon$) such that $K_\mathsf{S} \ge \eta K$.
As discussed before, we thus need to consider at least twice more blocks than outliers.
On the other hand, $K$ is by design upper bounded by $n$.
The global constraint can be written:
\begin{equation}\label{eq:constraint}
2n_\mathsf{O} = 2\varepsilon n < K \le n.
\end{equation}

Let $\alpha\colon [0, 1/2] \rightarrow [0, 1]$ such that: $\forall \varepsilon \in ]0, 1/2[$, $2\varepsilon < \alpha(\varepsilon) < 1$.
Several choices of acceptable function $\alpha$ are detailed in \Cref{tab:summary}, and illustrated in \Cref{fig:influence_apx}.
They include among others:
\begin{itemize}
\item the arithmetic mean: $\alpha(\varepsilon) = \frac{1 + 2\varepsilon}{2}$.
\item the geometric mean: $\alpha(\varepsilon) = \sqrt{2\varepsilon}$.
\item the harmonic mean: $\alpha(\varepsilon) = \frac{4\varepsilon}{1 + 2\varepsilon}$.
\item the polynomial: $\alpha(\varepsilon) = \varepsilon(5/2 - \varepsilon)$.
\end{itemize}

Once the function $\alpha$ is selected, \Cref{eq:constraint} is satisfied as soon as $K$ verifies:
\begin{equation*}
\alpha(\varepsilon) n \le K \le n.
\end{equation*}
It directly follows that
\begin{equation*}
K_\mathsf{S} = K - K_\mathsf{O} \ge K - n_\mathsf{O} \ge K - \varepsilon n \ge \left(1 - \frac{\varepsilon}{\alpha(\varepsilon)}\right)K = \frac{\alpha(\varepsilon) - \varepsilon}{\alpha(\varepsilon)}~K,
\end{equation*}
and one then may use
\begin{equation*}
\eta = \eta(\varepsilon) = \frac{\alpha(\varepsilon) - \varepsilon}{\alpha(\varepsilon)}.
\end{equation*}

Once $\eta(\varepsilon)$ is determined, a standard MoM deviation study can be carried out.
If at least $K/2$ sane blocks have an empirical estimate that is $t$ close to the expectation, then so is the MoM.
Reversing the implication gives:
\begin{align}
\mathbb{P}\Big\{\big|\hat{\theta}_\mathrm{MoM} - \theta\big| > t\Big\} &\le \mathbb{P}\left\{\sum_\text{blocks without outlier} \mathbbm{1}\Big\{ \big| \hat{\theta}_\text{block} - \theta\big| > t\Big\} \ge K_\mathsf{S} - \frac{K}{2}\right\},\nonumber\\
&\le \mathbb{P}\left\{\sum_\text{blocks without outlier} \mathbbm{1}\Big\{ \big| \hat{\theta}_\text{block} - \theta\big| > t\Big\} \ge \frac{2\eta(\varepsilon) - 1}{2\eta(\varepsilon)}K_\mathsf{S}\right\},\label{eq:dev_mom}
\end{align}
with $\hat{\theta}_\text{block} = (1/B) \sum_{i \in \text{block}} Z_i$ the block empirical mean.
Now observe that \Cref{eq:dev_mom} describes the deviation of a binomial random variable, with $K_\mathsf{S}$ trials and parameter $p_t = \mathbb{P}\{ |\hat{\theta}_\text{block} - \theta| > t\}$.
It can thus be upper bounded by
\begin{align*}
\sum_{k = \left\lceil\frac{2\eta(\varepsilon) - 1}{2\eta(\varepsilon)}K_\mathsf{S}\right\rceil}^{K_\mathsf{S}} \binom{K_\mathsf{S}}{k}~p_t^k~(1-p_t)^{K_\mathsf{S} - k} ~&\le~ p_t^{\frac{2\eta(\varepsilon) - 1}{2\eta(\varepsilon)}K_\mathsf{S}}~\sum_{k=1}^{K_\mathsf{S}}\binom{K_\mathsf{S}}{k},\\
&\le~ p_t^{\frac{2\eta(\varepsilon) - 1}{2\eta(\varepsilon)}K_\mathsf{S}}~2^{K_\mathsf{S}},\\[0.2cm]
&\le~ p_t^{\frac{2\eta(\varepsilon) - 1}{2}K}~2^{\eta(\varepsilon) K}.
\end{align*}

By virtue of Chebyshev's inequality, it holds that $p_t \le \sigma^2 / (Bt^2)$, with $B = \lfloor n/K \rfloor$ denoting the size of the blocks.
The right-hand side can then be rewritten as
\begin{equation*}
\exp\left(\frac{2\eta(\varepsilon) - 1}{2} K \cdot \log \left[2^{\frac{2\eta(\varepsilon)}{2\eta(\varepsilon) - 1}}\frac{\sigma^2}{Bt^2}\right]\right).
\end{equation*}
It can be set to $\delta$ by choosing $K = \left\lceil \frac{2}{2\eta(\varepsilon) - 1}\log(1/\delta)\right\rceil$, we will see later how this is compatible with the initial constraint $\alpha(\varepsilon) n \le K \le n$, and $t$ such that $2^{\frac{2\eta(\varepsilon)}{2\eta(\varepsilon) - 1}}\sigma^2/(Bt^2) = 1/e$, or again:
\begin{align}
t &= \sqrt{e}\sigma~\sqrt{\frac{2^{\frac{2\eta(\varepsilon)}{2\eta(\varepsilon) - 1}}}{B}},\nonumber\\
&\le \sqrt{e}\sigma~\sqrt{\frac{4\eta^2(\varepsilon)}{(2\eta(\varepsilon) - 1)^2}\frac{2 K}{n}},\nonumber\\
&\le 4\sqrt{e} \sigma~\frac{\eta(\varepsilon)}{(2\eta(\varepsilon) - 1)^\frac{3}{2}}~\sqrt{\frac{1 + \log(1/\delta)}{n}},\label{eq:final_bound}
\end{align}

where we have used $2^\frac{1}{x} \le 1/x^2$ for $x \le 1/2$, and $\lfloor x \rfloor \ge x/2$ for $x \ge 1$.

The final writing is obtained by setting
\begin{equation*}
\beta(\varepsilon) = \frac{2}{2\eta(\varepsilon) - 1} = \frac{2\alpha(\varepsilon)}{\alpha(\varepsilon) - 2\varepsilon},
\end{equation*}
and
\begin{equation*}
\gamma(\varepsilon) = \frac{\eta(\varepsilon)}{(2\eta(\varepsilon) - 1)^\frac{3}{2}} = \frac{\sqrt{\alpha(\varepsilon)}(\alpha(\varepsilon) - \varepsilon)}{(\alpha(\varepsilon) - 2\varepsilon)^{\frac{3}{2}}}.
\end{equation*}

Finally, the first part of the proof is achieved by ensuring that $K$ satisfies the initial constraint.
To do so, one may restrict the interval of acceptable $\delta$'s.
Indeed, it is enough for $\delta$ to satisfy:
\begin{gather*}
\alpha(\varepsilon) n ~\le~ \beta(\varepsilon)\log(1/\delta) ~\le~ n,\\[0.3cm]
e^{-n/\beta(\varepsilon)} ~\le~ \delta ~\le~ e^{-n\alpha(\varepsilon)/\beta(\varepsilon)}.\label{eq:interval}
\end{gather*}

The limitation on the range of $\delta$ is typical of MoM's concentration proofs.
The left limitation is due to the constraint $K \le n$, and is not very compelling in practice.
The right limitation comes from the constraint $ 2n_\mathsf{O} < K$ (or $\alpha(\varepsilon)n \le K$), and is specific to our outlier framework.
The purpose of the second part of \Cref{prop:mom} is precisely to remove the left limitation, under the assumption that $Z$ is $\rho$ sub-Gaussian.

Assume now that $Z$ is $\rho$ sub-Gaussian.
Chernoff's bound now gives that $p_t \le 2e^{-Bt^2/2\rho^2}$.
Plugging this bound into MoM's deviation yields
\begin{align*}
\mathbb{P}\Big\{\big|\hat{\theta}_\mathrm{MoM} - \theta\big| > t\Big\} &\le \exp\left(\frac{2\eta(\varepsilon) - 1}{2} K \cdot \log \left[2^{\frac{4\eta(\varepsilon) - 1}{2\eta(\varepsilon) - 1}} e^{-Bt^2/2\rho^2}\right]\right),\\
&\le \exp\left(-\frac{2\eta(\varepsilon) - 1}{16\rho^2} nt^2\right),
\end{align*}
for all $t$ such that
\begin{equation*}
t^2 \ge \frac{4 \rho^2}{B}~\frac{4\eta(\varepsilon) - 1}{2\eta(\varepsilon) - 1}\log 2,
\end{equation*}

Reverting in $\delta$ gives that it holds with probability at least $1 - \delta$
\begin{equation*}
\big|\hat{\theta}_\mathrm{MoM} - \theta\big| \le \frac{4\rho}{\sqrt{2\eta(\varepsilon) - 1}}~\sqrt{\frac{\log(1/\delta)}{n}},
\end{equation*}
for all $\delta$ that satisfies
\begin{equation}\label{eq:delta_inf}
\delta \le e^{-\frac{\log 2}{4}(4\eta(\varepsilon) - 1)\frac{n}{B}}, \text{\quad and in particular \quad} \delta \le e^{-4n \alpha(\varepsilon)}.
\end{equation}

Indeed it holds $B = \lfloor n/K \rfloor \ge n/(2K)$, so that $n/B \le 2K = 2 \lceil \alpha(\varepsilon) n\rceil \le 2(\alpha(\varepsilon)n + 1)\le 4\alpha(\varepsilon)n$, since $1 \le 2n_\mathsf{O} = 2\varepsilon n \le \alpha(\varepsilon)n$.
When $n_\mathsf{O} = \varepsilon = 0$, one may choose $K = 1$, $B=n$, and $\delta \le 1/e$.

The final writing is obtained by setting:
\begin{equation*}
\Gamma(\varepsilon) = \frac{1}{\sqrt{2\eta(\varepsilon) - 1}} = \sqrt{\frac{\alpha(\varepsilon)}{\alpha(\varepsilon) - 2\varepsilon}}.
\end{equation*}

To get the expectation bound, one may simply integrate the previously found deviation probabilities.
Reverting the inequality gives that it holds
\begin{equation*}
\mathbb{P}\left\{\big|\hat{\theta}_\mathrm{MoM} - \theta\big| > t\right\} \le e^{- \frac{nt^2}{16 \rho^2 \Gamma^2(\varepsilon)}},
\end{equation*}
for all $t$ such that (using \Cref{ass:sub_linear}):
\begin{equation}\label{eq:t_sup}
t \ge 8\rho~\Gamma(\varepsilon)\sqrt{\alpha(\varepsilon)}, \text{\quad and in particular \quad} t \ge 8\rho~\Gamma(\varepsilon)\sqrt{\frac{\alpha(\varepsilon)}{\varepsilon}}\frac{C_\mathsf{O}}{n^{(1 - \alpha_\mathsf{O})/2}}.
\end{equation}
One finally gets
\begin{align*}
\mathbb{E}\left[\big|\hat{\theta}_\mathrm{MoM} - \theta\big|\right] &= \int_0^\infty \mathbb{P}\left\{\big|\hat{\theta}_\mathrm{MoM} - \theta\big| > t\right\}dt,\\[0.3cm]
&\le \int_0^{8\rho~\Gamma(\varepsilon)\sqrt{\frac{\alpha(\varepsilon)}{\varepsilon}}\frac{C_\mathsf{O}}{n^{(1 - \alpha_\mathsf{O})/2}}} 1dt + \int_0^\infty e^{- \frac{nt^2}{16 \rho^2 \Gamma^2(\varepsilon)}}dt,\\[0.3cm]
&\le 8\rho~\Gamma(\varepsilon)\sqrt{\frac{\alpha(\varepsilon)}{\varepsilon}}\frac{C_\mathsf{O}}{n^{(1 - \alpha_\mathsf{O})/2}} + \frac{2\sqrt{\pi}\rho~\Gamma(\varepsilon)}{\sqrt{n}},\\[0.3cm]
&\le 2\rho~\Gamma(\varepsilon)\left(4C_\mathsf{O}~\frac{\Delta(\varepsilon)}{n^{(1 - \alpha_\mathsf{O})/2}} + \sqrt{\frac{\pi}{n}}\right),
\end{align*}
with the notation
\begin{equation*}
\Delta(\varepsilon) = \sqrt{\frac{\alpha(\varepsilon)}{\varepsilon}}.
\end{equation*}
\qed\bigskip

\begin{remark}\label{rmk:hoeffding}
Coming back to \Cref{eq:dev_mom}, one may also use Hoeffding's inequality to get:
\begin{align}\label{eq:hoef_pt}
\mathbb{P}\Big\{\big|\hat{\theta}_\mathrm{MoM} - \theta\big| > t\Big\} &\le \mathbb{P}\left\{\frac{1}{K_\mathsf{S}}\sum_\text{blocks without outlier} \mathbbm{1}\Big\{ \big| \hat{\theta}_\text{block} - \theta\big| > t\Big\} - p_t \ge \frac{2\eta(\varepsilon) - 1}{2\eta(\varepsilon)} - \frac{\sigma^2}{Bt^2}\right\},\nonumber\\
&\le \exp\left(-2\eta(\varepsilon) K\left(\frac{2\eta(\varepsilon) - 1}{2\eta(\varepsilon)} - \frac{\sigma^2}{Bt^2}\right)^2\right).
\end{align}
The right-hand side can be set to $\delta$ by choosing $K = \left\lceil \frac{9}{2} \frac{\eta(\varepsilon)}{(2\eta(\varepsilon) - 1)^2}\log(1/\delta)\right\rceil$, and $t$'s that satisfy:
\begin{align*}
\frac{2\eta(\varepsilon) - 1}{6\eta(\varepsilon)} &= \frac{\sigma^2}{Bt^2},\\
t &= \sqrt{6}\sigma~\sqrt{\frac{\eta(\varepsilon)}{2\eta(\varepsilon)-1}}~\frac{1}{\sqrt{B}},\\
t &\le \sqrt{6}\sigma~\sqrt{\frac{\eta(\varepsilon)}{2\eta(\varepsilon)-1}}~\sqrt{\frac{2K}{n}},\\
t &\le 3\sqrt{6}\sigma~\frac{\eta(\varepsilon)}{(2\eta(\varepsilon)-1)^{\frac{3}{2}}}~\sqrt{\frac{1 + \log(1/\delta)}{n}}.
\end{align*}
Up to the constant term which is bigger ($3\sqrt{6}$ instead of $4\sqrt{e}$), and the number of blocks which is more important, the latter result is very similar to \Cref{eq:final_bound}.
But constant factors were not the only reason motivating our choice of using the Binomial concentration.
Indeed, it should be noticed that the Hoeffding bound becomes vacuous when using $p_t \le 2\exp(-Bt^2/2\rho^2)$ for a $\rho$ sub-Gaussian r.v. $Z$.
Even if this sharper bound for $p_t$ is plugged in \Cref{eq:hoef_pt}, the quantity $(2\eta(\varepsilon) - 1)/(2\eta(\varepsilon)) - 2\exp(-Bt^2/(2\rho^2))$ may never go to $0$, making it impossible to improve the confidence range similarly to what has been done in \Cref{prop:mom}.
Notice that the same problem arises in the proof of \Cref{prop:mou_2sample}.
\end{remark}


\subsection{Proof of Proposition \ref{prop:mou_1sample}}
\label{apx:proof_mou_1_sample}

The proof of \Cref{prop:mom} can be fully reused, up to two details related to $U$-statistics.
The first one is Chebyshev's inequality, used to bound $p_t$ in the general case.
The latter now features the variance of the $U$-statistic, that can be upper bounded as follows.
Using the notation of \citet{van2000asymptotic} (see Chapter 12 therein), for $c \le d$ define $\zeta_c(h) = \mathrm{Cov}(h(Z_{i_1}, \ldots, Z_{i_d}),$ $h(Z_{i'_1}, \ldots, Z_{i'_d}))$ when $c$ variables are common.
Noticing that $\zeta_0(h) = 0$, it holds:
\begin{align*}
\mathrm{Var}\left(\bar{U}_B(h)\right) &= \mathrm{Cov}\left(\frac{1}{\binom{B}{d}} \sum_{i_1 < \ldots < i_d} h\left(Z_{i_1}, \ldots, Z_{i_d}\right), \frac{1}{\binom{B}{d}} \sum_{i'_1 < \ldots < i'_d} h\left(Z_{i'_1}, \ldots, Z_{i'_d}\right)\right),\\[0.3cm]
&= \frac{1}{\binom{B}{d}^2} \sum_{\substack{i_1 < \ldots < i_d\\i'_1 < \ldots < i'_d}} \mathrm{Cov}\left(h\left(Z_{i_1}, \ldots, Z_{i_d}\right), h\left(Z_{i'_1}, \ldots, Z_{i'_d}\right)\right),\\[0.3cm]
&= \frac{1}{\binom{B}{d}} \sum_{c=1}^d \binom{d}{c}\binom{B-d}{d-c} \zeta_c(h),\\[0.3cm]
&= \sum_{c=1}^d \frac{d!^2}{c! (d-c)!^2}~\frac{(B-d)(B-d-1)\ldots(B-2d+c+1)}{B(B-1)\ldots(B-d+1)}~\zeta_c(h),\\[0.3cm]
&\le d!~\frac{\sum_{c=1}^d \binom{d}{c}\zeta_c(h)}{B},\\
&= \frac{\Sigma^2(h)}{B},
\end{align*}
with $\Sigma^2(h) =  d! \sum_{c=1}^d \binom{d}{c}\zeta_c(h)$.

The second critical point that should be adapted is the upper bound $p_t \le 2e^{-Bt^2/2\rho^2}$ when $Z$ is $\rho$ sub-Gaussian.
If kernel $h$ is bounded, then Hoeffding's inequality for $U$-statistics \citep{Hoeffding63} gives instead that $p_t \le 2e^{-Bt^2/2d\|h\|_\infty^2}$.
The rest of the proof is similar to that of \Cref{prop:mom}.
We stress that Hoeffding's inequality is used on a sane block, so that we only need $h$ to be bounded if applied to r.v. $Z$.
In particular, it needs not be bounded on the outliers.
This happens \textit{e.g.} for any continuous kernel $h$ and r.v. $Z$ with bounded support.
\qed


\subsection{Proof of \Cref{prop:mou_2sample}}
\label{apx:cross_block}

Let us first recall the notation needed to the analysis of $\hat{\theta}_{\mathrm{MoU}_2}(H)$.
The numbers of blocks are denoted by $K_X$ and $K_Y$, and the block sizes by $B_X = \lfloor n/K_X \rfloor$ and $B_Y = \lfloor m/K_Y \rfloor$ respectively.
The number of sane blocks are denoted by $K_{X, \mathsf{S}}$ and $K_{Y, \mathsf{S}}$, and for $k \le K_X$ and $l \le K_Y$, we set:
\begin{equation*}
\hat{U}_{k,l}(H) = \frac{1}{B_X B_Y} \sum_{i \in \mathcal{B}^X_k} \sum_{j \in \mathcal{B}^Y_l} H(X_i, Y_j),
\end{equation*}
the (two-sample) $U$-statistic built upon blocks $\mathcal{B}^X_k$ and $\mathcal{B}^Y_l$.
Let $I_{k,l}^t = \mathbbm{1}\{|\hat{U}_{k,l}(H) - \theta(H)| > t\}$ be the indicator random variable characterizing its $t$-closeness to the true parameter $\theta(H)$.

As previously discussed, the constraint on $K_X$ and $K_Y$ now writes:
\begin{equation}\label{eq:constraint2}
\alpha(\varepsilon_X + \varepsilon_Y - \varepsilon_X\varepsilon_Y) nm \le K_X K_Y \le nm.
\end{equation}
In order to simplify the computation, we will however consider the following double constraint:
\begin{equation}\label{eq:constraint_decouple}
\left\{\begin{matrix}
\sqrt{\alpha(\varepsilon_X + \varepsilon_Y - \varepsilon_X\varepsilon_Y)}n \le K_X \le n,\\[0.4cm]
\sqrt{\alpha(\varepsilon_X + \varepsilon_Y - \varepsilon_X\varepsilon_Y)}m \le K_Y \le m.
\end{matrix}\right.
\end{equation}

\Cref{eq:constraint_decouple} naturally implies \Cref{eq:constraint2}, and one may observe that it does not impact the limit condition $\varepsilon_X + \varepsilon_Y - \varepsilon_X\varepsilon_Y < 1/2$.
Similarly to previous proofs, \Cref{eq:constraint2} yields
\begin{equation*}
K_{X, \mathsf{S}}K_{Y, \mathsf{S}} \ge\left(1 - \frac{\varepsilon_X + \varepsilon_Y - \varepsilon_X\varepsilon_Y}{\alpha(\varepsilon_X + \varepsilon_Y - \varepsilon_X\varepsilon_Y)}\right) K_XK_Y \coloneqq \eta_{XY} \cdot K_X K_Y,
\end{equation*}
for notation simplicity.
On the other hand, \Cref{eq:constraint_decouple} ensures both
\begin{equation*}
\left\{\begin{matrix}
K_{X, \mathsf{S}} \ge \left(1 - \frac{\varepsilon_X}{\sqrt{\alpha(\varepsilon_X + \varepsilon_Y - \varepsilon_X\varepsilon_Y)}}\right)K_X \coloneqq \eta_X \cdot K_X,\\[0.4cm]
K_{Y, \mathsf{S}} \ge \left(1 - \frac{\varepsilon_Y}{\sqrt{\alpha(\varepsilon_X + \varepsilon_Y - \varepsilon_X\varepsilon_Y)}}\right)K_Y \coloneqq \eta_Y \cdot K_Y,
\end{matrix}\right.
\end{equation*}
with a slight abuse of notation since $\eta_X$ also depends on $Y$ (and conversely).
Notice that it holds true $1/2 \le \eta_X, \eta_Y \le 1$.
Using the same reasoning as before, one gets:
\begin{align*}
\mathbb{P}\Big\{\big|\hat{\theta}_{\mathrm{MoU}_2}(H) - \theta(H)\big| > t\Big\} \le~& \mathbb{P}\left\{\sum_{k=1}^{K_X} \sum_{l=1}^{K_Y} I_{k,l}^t \ge \frac{K_X K_Y}{2}\right\},\\
\le~& \mathbb{P}\left\{\underset{\text{blocks without outlier}}{\sum\sum} I_{k,l}^t \ge \frac{2\eta_{XY} - 1}{2 \eta_{XY}}K_{X, \mathsf{S}} K_{Y, \mathsf{S}}\right\}.
\end{align*}

However, unlike \Cref{eq:dev_mom}, the above equation does not relate to a binomial random variable, as the $I_{k, l}^t$ are not independent, see \Cref{fig:mou_2}.
An elegant alternative then consists in leveraging the independence between samples $X$ and $Y$ and using Hoeffding's inequality.
\Cref{eq:var_ustat_multi} gives $\sigma^2_{B_X, B_Y}(H) \le \Sigma^2(H) / (B_X \wedge B_Y)$, with $\Sigma^2(H) = \sigma^2(H) + \sigma_1^2(H) + \sigma_2^2(H)$, so that:

\begin{align*}
\le~& \mathbb{P}\Bigg\{\frac{1}{K_{X, \mathsf{S}} K_{Y, \mathsf{S}}} \underset{\text{blocks w/o outlier}}{\sum\sum} I_{k, l}^t - \mathbb{E}\left[I_{k, l}^t \mid \bm{X}\right] + \mathbb{E}\left[I_{k, l}^t \mid \bm{X}\right] - \mathbb{E}\left[I_{k, l}^t\right]\\
&\hspace{8cm}\ge \frac{2\eta_{XY} - 1}{2 \eta_{XY}} - \frac{\Sigma^2(H)}{(B_X \wedge B_Y)t^2}\Bigg\},\\
\le~& \mathbb{P}\left\{\frac{1}{K_{Y, \mathsf{S}}} \sum_{l=1}^{K_{Y, \mathsf{S}}} J_l^t - \mathbb{E}\left[J_l^t \mid \bm{X}\right] \ge \frac{2\eta_{XY} - 1}{4 \eta_{XY}} - \frac{\Sigma^2(H)}{2(B_X \wedge B_Y)t^2}\right\}+\\
& \mathbb{P}\left\{\frac{1}{K_{X, \mathsf{S}}}\sum_{k=1}^{K_{X, \mathsf{S}}} \mathbb{E}\left[I_{k, l}^t \mid \bm{X}\right] - \mathbb{E}\left[I_{k, l}^t\right] \ge \frac{2\eta_{XY} - 1}{4 \eta_{XY}} - \frac{\Sigma^2(H)}{2(B_X \wedge B_Y)t^2}\right\},\\
\le~& \exp\left(-2\eta_Y K_Y\left(\frac{2\eta_{XY} - 1}{4 \eta_{XY}} - \frac{\Sigma^2(H)}{2(B_X \wedge B_Y)t^2}\right)^2\right) +\\
&\exp\left(-2\eta_X K_X\left(\frac{2\eta_{XY} - 1}{4 \eta_{XY}} - \frac{\Sigma^2(H)}{2(B_X \wedge B_Y)t^2}\right)^2\right),
\end{align*}
with the notation $\displaystyle J_l^t = \frac{1}{K_{X, \mathsf{S}}} \sum_{k=1}^{K_{X, \mathsf{S}}} I_{k, l}^t$, and $\bm{X} = (X_1, \ldots, X_n)$.

Now the right-hand side is set to $\delta$ by choosing $K_Z = \left\lceil \frac{18~\eta_{XY}^2}{\eta_Z (2\eta_{XY} - 1)^2} \log(2/\delta)\right\rceil$ for $Z = X, Y$ respectively, and for $t$ that satisfies:
\begin{align*}
\frac{\Sigma^2(H)}{2(B_X \wedge B_Y)t^2} &= \frac{2\eta_{XY} - 1}{12 \eta_{XY}},\\[0.3cm]
t &= \Sigma(H)~\sqrt{\frac{6 \eta_{XY}}{2\eta_{XY} - 1}}~\sqrt{\frac{1}{B_X \wedge B_Y}},\\[0.3cm]
&\le \Sigma(H)~\sqrt{\frac{6 \eta_{XY}}{2\eta_{XY} - 1}}~\sqrt{\frac{2 \max(K_X, K_Y)}{n \wedge m}},\\[0.3cm]
&\le 12\sqrt{3}~\Sigma(H)\left(\frac{\eta_{XY}}{2\eta_{XY} - 1}\right)^\frac{3}{2}\sqrt{\frac{1 + \log(2/\delta)}{n \wedge m}},\\[0.3cm]
&\le 12\sqrt{3}~\Sigma(H)~\gamma(\varepsilon_X + \varepsilon_Y - \varepsilon_X\varepsilon_Y)~\sqrt{\frac{1 + \log(2/\delta)}{n \wedge m}}.
\end{align*}

Constraints \eqref{eq:constraint_decouple} are finally fulfilled by choosing $\delta$ such that:
\begin{gather*}
\left\{\begin{matrix}
\sqrt{\alpha(\varepsilon_X + \varepsilon_Y - \varepsilon_X\varepsilon_Y)}n \le \frac{18~\eta_{XY}^2}{\eta_X (2\eta_{XY} - 1)^2} \log(2/\delta) \le n,\\[0.4cm]
\sqrt{\alpha(\varepsilon_X + \varepsilon_Y - \varepsilon_X\varepsilon_Y)}m \le \frac{18~\eta_{XY}^2}{\eta_Y (2\eta_{XY} - 1)^2} \log(2/\delta) \le m,
\end{matrix}\right.\\[0.3cm]
2\max\left(e^{-n\beta_X}, e^{-m\beta_Y}\right) ~\le~ \delta ~\le~ 2\min\left(e^{-n\sqrt{\alpha}/\beta_X}, e^{-m\sqrt{\alpha}/\beta_Y}\right),
\end{gather*}
with the shortcut notation $\alpha = \alpha(\varepsilon_X + \varepsilon_Y - \varepsilon_X\varepsilon_Y)$, and $\beta_Z = \frac{18~\eta_{XY}^2}{\eta_Z (2\eta_{XY} - 1)^2}$ for $Z=X,Y$.
\qed


\subsection{Proof of \Cref{prop:mou_diag}}
\label{apx:mou_diag}

Again, the proof can be directly adapted from that of \Cref{prop:mom}.
The first difference lies in the constraint $K$ needs to satisfy.
It now writes: $2(n_\mathsf{O} + m_\mathsf{O}) = 2(\varepsilon_X + \varepsilon_Y)n < K \le n$, and the reasoning can then be reused in totality with $\varepsilon_X + \varepsilon_Y$ instead of $\varepsilon$.
The second difference is Chebyshev's inequality, but \Cref{eq:var_ustat_multi} gives that $\sigma^2_{B_X, B_Y}(H) \le \Sigma^2(H) / B$, with $\Sigma^2(H) = \sigma^2(H) + \sigma_1^2(H) + \sigma_2^2(H)$.
Finally, when $\|H\|_\infty$ is finite, using the notation $\bm{X} = (X_1, \ldots, X_n)$, one may bound $p_t$ as follows:
\begin{align*}
p_t =~& \mathbb{P}\left\{|\hat{U}_{1, 1}(H) - \theta(H)| > t\right\},\\[0.3cm]
=~& \mathbb{P}\bigg\{\Big|\frac{1}{B^2} \sum_{i \in \mathcal{B}^X_1}\sum_{j \in \mathcal{B}^Y_1} H(X_i, Y_j) - \theta(H)\Big| > t\bigg\},\\[0.3cm]
\le~&\mathbb{P}\bigg\{\bigg|\frac{1}{B} \sum_{j \in \mathcal{B}^Y_1}\bigg(\sum_{i \in \mathcal{B}^X_1} \frac{H(X_i, Y_j)}{B} - \mathbb{E}\bigg[\sum_{i \in \mathcal{B}^X_1} \frac{H(X_i, Y_j)}{B} ~\Big|~ \bm{X}\bigg]\bigg)\bigg| > \frac{t}{2} ~\Big|~ \bm{X} \bigg\}\\[0.3cm]
+~&\mathbb{P}\bigg\{\Big| \frac{1}{B}\sum_{i \in \mathcal{B}^X_1} \mathbb{E}_Y\big[H(X_i, Y)\big] - \theta(H)\Big| > \frac{t}{2} \bigg\},\\[0.3cm]
\le~& 2e^{-Bt^2/8\|H\|_\infty^2} + 2e^{-Bt^2/8\|H\|_\infty^2},
\end{align*}
where we have used Hoeffding's inequality twice: on the $\sum_{i \in \mathcal{B}^X_1} \frac{H(X_i, Y_j)}{B}$ for $j \in \mathcal{B}^Y_1$, conditionally to the $X_i$'s, and a second time to the $\mathbb{E}_Y \big[H(X_i, Y)\big]$ for $i \in \mathcal{B}^X_1$, both random variables being bounded by $\|H\|_\infty$.
The rest of the proof is similar to that of \Cref{prop:mom}.
\qed


\subsection{Extension to $U$-statistics of Arbitrary Degrees and Number of Samples}
\label{apx:ext_ustat}

Similarly to the extension from \Cref{prop:mom} to \Cref{prop:mou_1sample}, the first important step consists in upper bounding the variance of the $U$-statistic.
To allow an effective use of Chebyshev's inequality, the latter must be of the order $\mathcal{O}(1/n)$, where we recall that $n$ is the number of observations in the sample (or the size of the smallest sample in the case of a multisample $U$-statistic).
This is for instance the case in \Cref{eq:var_ustat_multi}, \textit{i.e.} for the $2$-sample $U$-statistic of degree $(1, 1)$.
As a first go, we detail here the derivation of \Cref{eq:var_ustat_multi}.
We then show that with similar computations, it is direct to show that for any $p$-sample $U$-statistic of degrees $(1, \ldots, 1)$, the $\mathcal{O}(1/n)$ condition holds.
Finally, we extend it to arbitrary degrees.
Recall that we compute the variance of the $2$-sample $U$-statistic of degrees $(1, 1)$, based on the samples $\mathcal{S}_n^X = \{X_1, \ldots, X_n\}$, and $\mathcal{S}_m^Y = \{Y_1, \ldots, Y_m\}$.
It holds:
\begin{align*}
\text{Var}\Big(\frac{1}{nm} &\sum_{i=1}^n \sum_{j=1}^m H(X_i, Y_j)\Big)\\
&= \frac{1}{n^2 m^2} \text{Var}\left(\sum_{i=1}^n \sum_{j=1}^m H(X_i, Y_j)\right),\\
&= \frac{1}{n^2 m^2} \mathbb{E}\left[\sum_{i,i'=1}^n \sum_{j,j'=1}^m H(X_i, Y_j) H(X_{i'}, Y_{j'})\right] - \theta^2(H),\\
&=\frac{1}{nm}\mathbb{E}\left[H^2(X,Y)\right] + \frac{m-1}{nm}\mathbb{E}\left[H(X,Y)H(X,Y')\right] + \frac{n-1}{nm} \mathbb{E}\left[H(X,Y)H(X',Y)\right] - \frac{n+m-1}{nm} \theta^2(H),\\[0.2cm]
&=\frac{1}{nm}\sigma^2(H) + \frac{m-1}{nm}\sigma^2_1(H) + \frac{n-1}{nm} \sigma^2_2(H),\\[0.2cm]
&\le \frac{\Sigma^2(H)}{n \wedge m},
\end{align*}
with $\Sigma^2(H) = \sigma^2(H) + \sigma_1^2(H) + \sigma_2^2(H)$, $\sigma^2(H) = \text{Var}\left(H(X,Y)\right)$, $\sigma^2_1(h) = \text{Cov}\left(H(X,Y), H(X,Y')\right) = \mathrm{Var}(H_1(X))$, with $H_1(x) = \mathbb{E}\left[H(x, Y)\right]$, and $\sigma^2_2(h) = \text{Cov}\left(H(X,Y), H(X',Y)\right) = \mathrm{Var}(H_2(Y))$, with $H_2(y) = \mathbb{E}\left[H(X, y)\right]$.

To highlight the mechanism at stake, we reproduce the above computations for a $3$-sample $U$-statistic of degrees $(1, 1, 1)$.
It is then direct to see that for any $p$-sample $U$-statistic of degrees $(1, \ldots, 1)$, the $\mathcal{O}(1/n)$ condition holds.
We have now at disposal a new sample $\mathcal{S}_q^Z = \{Z_1, \ldots, Z_q\}$, and the variance of the $U$-statistic writes:
\begin{align}
\text{Var}&\Big(\frac{1}{nmq} \sum_{i=1}^n \sum_{j=1}^m \sum_{k=1}^q H(X_i, Y_j, Z_k)\Big)\nonumber\\
&= \frac{1}{n^2 m^2 q^2} \text{Var}\left(\sum_{i=1}^n \sum_{j=1}^m \sum_{k=1}^q H(X_i, Y_j, Z_k)\right),\nonumber\\
&= \frac{1}{n^2 m^2 q^2} \mathbb{E}\left[\sum_{i,i'=1}^n \sum_{j,j'=1}^m \sum_{k, k'} H(X_i, Y_j, Z_k) H(X_{i'}, Y_{j'}, Z_{k'})\right] - \theta^2(H),\label{eq:covariance}\\[0.2cm]
&=\frac{1}{nmq}\mathbb{E}\left[H^2(X,Y,Z)\right] + \frac{(m-1)(q-1)}{nmq}\mathbb{E}\left[H(X,Y,Z)H(X,Y',Z')\right]\nonumber\\
&~~+ \frac{(n-1)(q-1)}{nmq}\mathbb{E}\left[H(X,Y,Z)H(X',Y,Z')\right] + \frac{(n-1)(m-1)}{nmq}\mathbb{E}\left[H(X,Y,Z)H(X',Y',Z)\right]\nonumber\\
&~~+ \frac{n-1}{nmq} \mathbb{E}\left[H(X,Y,Z)H(X',Y,Z)\right] + \frac{m-1}{nmq} \mathbb{E}\left[H(X,Y,Z)H(X,Y',Z)\right] + \frac{q-1}{nmq} \mathbb{E}\left[H(X,Y,Z)H(X,Y,Z')\right]\nonumber\\
&~~- \frac{nmq - (n-1)(m-1)(q-1)}{nmq}\theta^2(H),\nonumber\\[0.2cm]
&=\frac{1}{nmq}\sigma^2(H) + \frac{(m-1)(q-1)}{nmq}\sigma^2_1(H) + \frac{(n-1)(q-1)}{nmq}\sigma^2_2(H) + \frac{(n-1)(m-1)}{nmq}\sigma^2_3(H)\nonumber\\
&~~+ \frac{n-1}{nmq} \sigma^2_{23}(H) + \frac{m-1}{nmq} \sigma^2_{13}(H) + \frac{q-1}{nmq} \sigma^2_{12}(H)\nonumber\\[0.2cm]
&\le \frac{\Sigma^2(H)}{n \wedge m \wedge q},\nonumber
\end{align}

with $\Sigma^2(H) = \sigma^2(H) + \sigma_1^2(H) + \sigma_2^2(H) + \sigma_3^2(H) + \sigma_{23}^2(H) + \sigma_{13}^2(H) + \sigma_{12}^2(H)$, and with a notation abuse $\sigma_{i/ij}^2 = \mathrm{Var}\big(H_{i/ij}(X, Y, Z)\big)$, with $H_{i/ij}(X_1, X_2, X_3) = \mathbb{E}[H(X_1, X_2, X_3) \mid X_i]$ or $\mathbb{E}[H(X_1, X_2, X_3) \mid X_i, X_j]$ respectively.

From this second example we can extrapolate the mechanism that generates the variance of the $U$-statistic.
Coming back to \Cref{eq:covariance}, we have to compute a certain number of covariance terms.
The important thing that distinguishes the different covariances is the number of variables shared between $H(X_i, Y_j, Z_k)$ and $H(X_{i'}, Y_{j'}, Z_{k'})$.
Depending on this number, and on which variable(s) is (are) shared, one of the $\sigma^2_{i/ij}$ variances appears.
This variance is multiplied by the number of times a suitable combination arise.
For a shared variable, this is $n$ (respectively, $m$ or $q$, \textit{i.e.} the size of the associated sample).
For non-shared variables, this is $n(n-1)$.
As at least one variable is shared (otherwise the two terms are independent, and the expectation is then equal to $\theta^2(H)$, that cancels with the last term of \Cref{eq:covariance}), we end up with variance terms, multiplied by $1 / n_\text{min}$ at most (because of the $1/(n^2m^2q^2)$ factor).
This reasoning validates the $\mathcal{O}(1/n)$ condition discussed earlier, and is applicable to an arbitrary number of samples.
Notice finally that it can be shown that all partial variance terms are smaller than $\sigma^2(H) = \mathrm{Var}\big(H(X_1, \ldots, X_p)\big)$, so that a simple condition for all the variance terms to be finite is $\sigma^2(H) < +\infty$.
The same analysis also applies to arbitrary numbers of samples \textbf{and} degrees.
Combining it to the variance computation of \cref{apx:proof_mou_1_sample}, it is direct to show that the $\mathcal{O}(1/n)$ remains valid in this setting.\\

The second important step is the generalization of Hoeffding's inequality when the essential supremum is bounded.
There is no particular difficulty here, since Hoeffding's inequality for $U$-statistics of arbitrary degrees can be used, possibly combined with the condition trick introduced in the previous section when several samples are considered.


\subsection{Proof of \Cref{thm:mou_min}}
\label{apx:rade}

Using the fact that $\hat{g}_\mathrm{MoU}$ minimizes $\mathrm{MoU}_{\mathcal{S}_n}(\ell_g)$ over $\mathcal{G}$, one gets:\vspace{0.2cm}
\begin{align*}
\mathcal{R}(\hat{g}_\mathrm{MoU}) - \mathcal{R}(g^*) &\le \mathcal{R}(\hat{g}_\mathrm{MoU}) - \mathrm{MoU}_{\mathcal{S}_n}(\ell_{\hat{g}_\mathrm{MoU}}) + \mathrm{MoU}_{\mathcal{S}_n}(\ell_{g^*}) - \mathcal{R}(g^*),\\[0.2cm]
&\le 2 \sup_{g \in \mathcal{G}} \big|\mathrm{MoU}_{\mathcal{S}_n}(\ell_g) - \mathcal{R}(g)\big|,\\
&\le 2 \sup_{g \in \mathcal{G}} \left|\mathrm{MoU}_{\mathcal{S}_n}(\ell_g) - \mathbb{E}[\ell_g]\right|.
\end{align*}

For a fixed $g \in \mathcal{G}$, \Cref{prop:mou_1sample} and \Cref{hyp:bounded_loss} gives that for all $\delta \in ]0, \exp(-4n\alpha(\varepsilon))]$, we have with probability larger than $1 - \delta$:
\begin{equation*}
\big|\mathrm{MoU}_{\mathcal{S}_n}(\ell_g) - \mathbb{E}[\ell_g]\big| ~\le~ 4 \sqrt{2}M~\Gamma(\varepsilon)~\sqrt{\frac{\log(1/\delta)}{n}}.
\end{equation*}

By virtue of Sauer's lemma, \Cref{hyp:vc_dim} altogether with the union bound then gives that for all $\delta \in ]0, \exp(-4\Delta^2(\varepsilon)n_\mathsf{O})]$, it holds with probability at least $1 - \delta$:
\begin{equation*}
\sup_{g \in \mathcal{G}}\big|\mathrm{MoU}_{\mathcal{S}_n}(\ell_g) - \mathbb{E}[\ell_g]\big| ~\le~ 4 \sqrt{2}M~\Gamma(\varepsilon)~\sqrt{\frac{\textsc{VC}_\text{dim}(\mathcal{G})(1 + \log(n)) +\log(1/\delta)}{n}}.
\end{equation*}
\qed


\subsection{Generalization Bound via Entropic Complexity}
\label{apx:chaining}

In this section, we highlight the versatility of the concentration bounds established in \Cref{sec:revisit} by deriving generalization guarantees through another complexity assumption than that used in \Cref{thm:mou_min}.
Namely, we use the following entropic characterization.
\begin{assumption}
The collection of functions $\mathcal{L}_\mathcal{G}=\{\ell_g \colon g \in \mathcal{G}\}$ is a uniform Donsker class (relative to $\|\cdot\|_\infty$) with polynomial uniform covering numbers, \textit{i.e.} there exist constants $C_\mathcal{G}>0$ and $r\geq 1$ such that: $\forall \zeta>0$,
$$
\mathcal{N}(\zeta, \mathcal{L}_\mathcal{G}, L_\infty(Q) )\leq C_\mathcal{G}(1/\zeta)^r,
$$
where $\mathcal{N}(\zeta, \mathcal{L}_\mathcal{G}, \|\cdot\|_\infty)$ denotes the number of $\|\cdot\|_\infty$-balls of radius $\zeta>0$ needed to cover class $\mathcal{L}_\mathcal{G}$.
\end{assumption}

Now, let $\zeta > 0$, and $\ell_1, \ldots, \ell_{\mathcal{N}(\zeta, \mathcal{L}_\mathcal{G}, \|\cdot\|_\infty)}$ be a $\zeta$-coverage of $\mathcal{L}_\mathcal{G}$ with respect to $\|\cdot\|_\infty$.
From now on, we use $\mathcal{N}=\mathcal{N}(\zeta, \mathcal{L}_\mathcal{G}, \|\cdot\|_\infty)$ for notation simplicity.
Let $\ell_g$ be an arbitrary element of $\mathcal{L}_\mathcal{G}$.
By definition, there exists $i \le \mathcal{N}$ such that $\|\ell_g - \ell_i\|_\infty \le \zeta$.
It holds then:
\begin{align}
\big|\mathrm{MoU}_{\mathcal{S}_n}(\ell_g) - \mathbb{E}\left[\ell_g\right]\big| &\le \big|\mathrm{MoU}_{\mathcal{S}_n}(\ell_g) - \mathrm{MoU}_{\mathcal{S}_n}(\ell_i)\big| + \big|\mathrm{MoU}_{\mathcal{S}_n}(\ell_i) - \mathbb{E}\left[\ell_i\right]\big| + \big|\mathbb{E}\left[\ell_i\right] - \mathbb{E}\left[\ell_g\right]\big|,\nonumber\\
&\le 2\zeta + \big|\mathrm{MoU}_{\mathcal{S}_n}(\ell_i) - \mathbb{E}\left[\ell_i\right]\big|.\label{eq:chain}
\end{align}

Applying the second claim of \Cref{prop:mou_1sample} to every $\ell_i$, the union bound gives that for all $\delta \in ]0, e^{-4n\alpha(\varepsilon)}]$, choosing $K = \lceil \alpha(\varepsilon)n\rceil$, it holds with probability at least $1 - \delta$:
\begin{equation*}
\sup_{i \le \mathcal{N}} \big|\mathrm{MoU}_{\mathcal{S}_n}[\ell_i] - \mathbb{E}\left[\ell_i\right]\big| \le 4\sqrt{2}M\Gamma(\varepsilon)\sqrt{\frac{\log(\mathcal{N}/\delta)}{n}}.
\end{equation*}

Taking the supremum in both sides of \Cref{eq:chain}, it holds with probability at least $1 - \delta$:
\begin{equation*}
\sup_{g \in \mathcal{G}} \big|\mathrm{MoU}_{\mathcal{S}_n}[\ell_g] - \mathbb{E}\left[\ell_g\right]\big| \le 2\zeta + 4\sqrt{2}M\Gamma(\varepsilon)\sqrt{\frac{\log(\mathcal{N}/\delta)}{n}}.
\end{equation*}

Choosing $\zeta \sim 1/ \sqrt{n}$, it holds with probability at least $1 - \delta$:
\begin{equation*}
\sup_{g \in \mathcal{G}} \big|\mathrm{MoU}_{\mathcal{S}_n}[\ell_g] - \mathbb{E}\left[\ell_g\right]\big| \le \frac{2}{\sqrt{n}} + 4\sqrt{2}M\Gamma(\varepsilon)\sqrt{\frac{(r/2)\log(n) + \log(C_\mathcal{G}/\delta)}{n}}.
\end{equation*}
We recover the bound of \Cref{thm:mou_min} up to a $\log(n)$ factor.


\subsection{Proof of \Cref{thm:alg}}
\label{apx:thm_alg}

First, we detail the assumptions needed to derive \Cref{thm:alg}, that were not explicited in the core text due to space constraints.
They are adaptations of the Assumptions used to derive Theorem 3 in \citet{lecue2018robust}.
They state as follows.
\begin{itemize}
\item for any $u \in \mathbb{R}^p$ and $z, z' \in \mathcal{Z}^2$, it holds: $\big\|\nabla_u \ell(g_u, z, z')\big\| \le L$,
\item for any sample $\mathcal{S}_n$, there exists a unique minimum $u_\text{min} = \argmin_{u \in \mathbb{R}^p} \mathbb{E}_\text{part}\left[\mathrm{MoU}_{\mathcal{S}_n}(\ell_g) \mid \mathcal{S}_n\right]$, where the expectation is taken with respect to all possible ways of partitioning of sample $\mathcal{S}_n$,
\item $\sum_{t=1}^\infty \gamma_t = +\infty$, \text{ and } $\sum_{t=1}^\infty \gamma_t^2 < +\infty$,
\item for any sample $\mathcal{S}_n$, model $u \in \mathbb{R}^p$, and $\epsilon > 0$, it holds: $\inf_{\|u - u_\text{min}\| > \epsilon} (u - u_\text{min})^\top \mathbb{E}_\text{part}\left[\nabla_u\mathrm{MoU}_{\mathcal{S}_n}(\ell_g) \mid \mathcal{S}_n\right] < 0$,
\item for any sample $\mathcal{S}_n$ and model $u \in \mathbb{R}^p$, there exists an open convex set $\mathscr{B}$ containing $u$ such that for any equipartition of $\{1, \ldots, N\}$ into $K$ blocks $\mathcal{B}_1, \ldots, \mathcal{B}_k$
there exists $k_\text{med} \le K$ such that for all $v \in \mathscr{B}$, $\mathcal{B}_{k_\text{med}}$ is the median block.
\end{itemize}

Under these five assumptions, a direct adaptation of Theorem 3 in \citet{lecue2018robust} then gives the almost sure convergence of the output of \Cref{alg:gd} towards $u_\text{min}$.
We have now to study the excess risk of $\hat{g}_\text{alg} = g_{u_\text{min}}$.
Jensen's inequality gives:
\begin{equation*}
\mathcal{R}(\hat{g}_\mathrm{alg}) - \mathcal{R}(g^*) \le 2 \sup_{g \in \mathcal{G}} \big|\mathbb{E}_\text{part}\left[\mathrm{MoU}_{\mathcal{S}_n}(\ell_g)\right] - \mathcal{R}(g)\big| \le 2~\mathbb{E}_\text{part}\Big[\sup_{g \in \mathcal{G}} \big|\mathrm{MoU}_{\mathcal{S}_n}(\ell_g) - \mathbb{E}[\ell_g]\big|\Big].
\end{equation*}
Applying \Cref{thm:mou_min} then allows to upper bound the right-hand side with high probability, and to conclude.
\qed

\section{Numerical Experiments}
\label{apx:expes}

In this section, we present numerical experiments highlighting the remarkable \emph{robustness-to-outliers} of MoM-based estimators.
In particular, we present mean and (multisample) $U$-statistics estimation experiments under \Cref{ass:sub_linear}, that emphasize the superiority of MoM/MoU/MoU$_2$ compared to standard alternatives (see \Cref{exp:esti}).
We also provide implementations of \Cref{alg:gd} on both ranking and metric learning problems (\Cref{exp:learning}).
They illustrate the good behavior of the MoU Gradient Descent (MoU-GD) when the training dataset is contaminated.

\subsection{Estimation Experiments}
\label{exp:esti}

For all our experiments, we set $n_\mathsf{O} = \sqrt{n}$, so that \Cref{ass:sub_linear} is fulfilled with $C_\mathsf{O} = 1$, $\alpha_\mathsf{O} = 1/2$.
We next specify particular instances of \Cref{hyp:framework}, \textit{i.e.} a distribution for $Z$ (or for $X$ and $Y$), and a distribution for the outliers, such that standard estimators are dramatically damaged, while the MoM-based versions studied in the present article are barely impacted, corroborating the theoretical guarantees established in \Cref{prop:mom,prop:mou_1sample,prop:mou_diag}.
We have selected $K$ according to the Harmonic upper bound, so that \Cref{hyp:config} is fulfilled as well.

\begin{figure}[!t]
\begin{center}
\begin{subfigure}{0.45\textwidth}
\includegraphics[width=\textwidth]{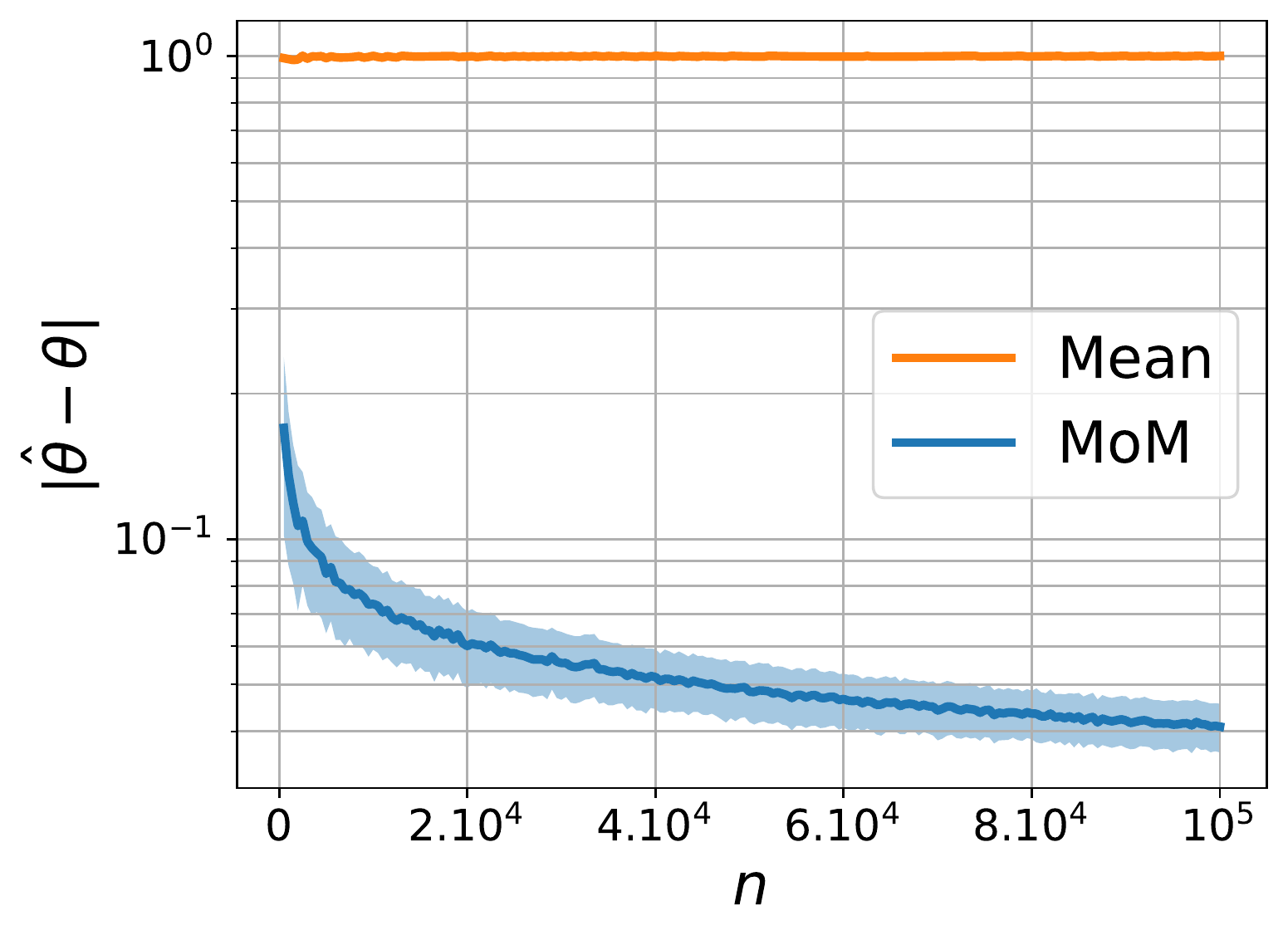}
\caption{Ruining the mean (avg. 500 runs).}
\label{fig:break_mean}
\end{subfigure}
\hfill
\begin{subfigure}{0.45\textwidth}
\includegraphics[width=\textwidth]{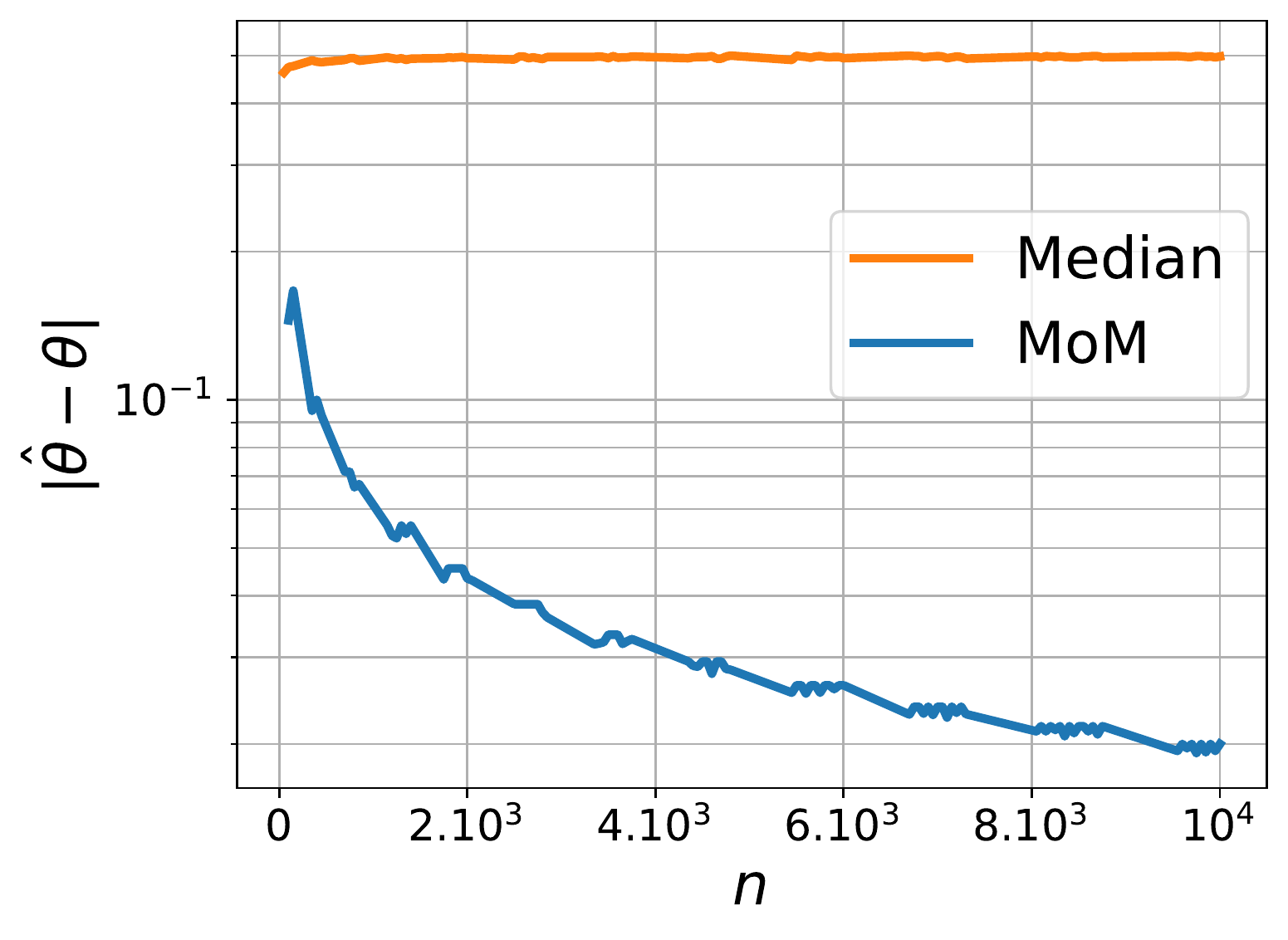}
\caption{Ruining the median (avg. 500 runs).}
\label{fig:break_median}
\end{subfigure}\\[0.5cm]
\begin{subfigure}{0.45\textwidth}
\includegraphics[width=\textwidth]{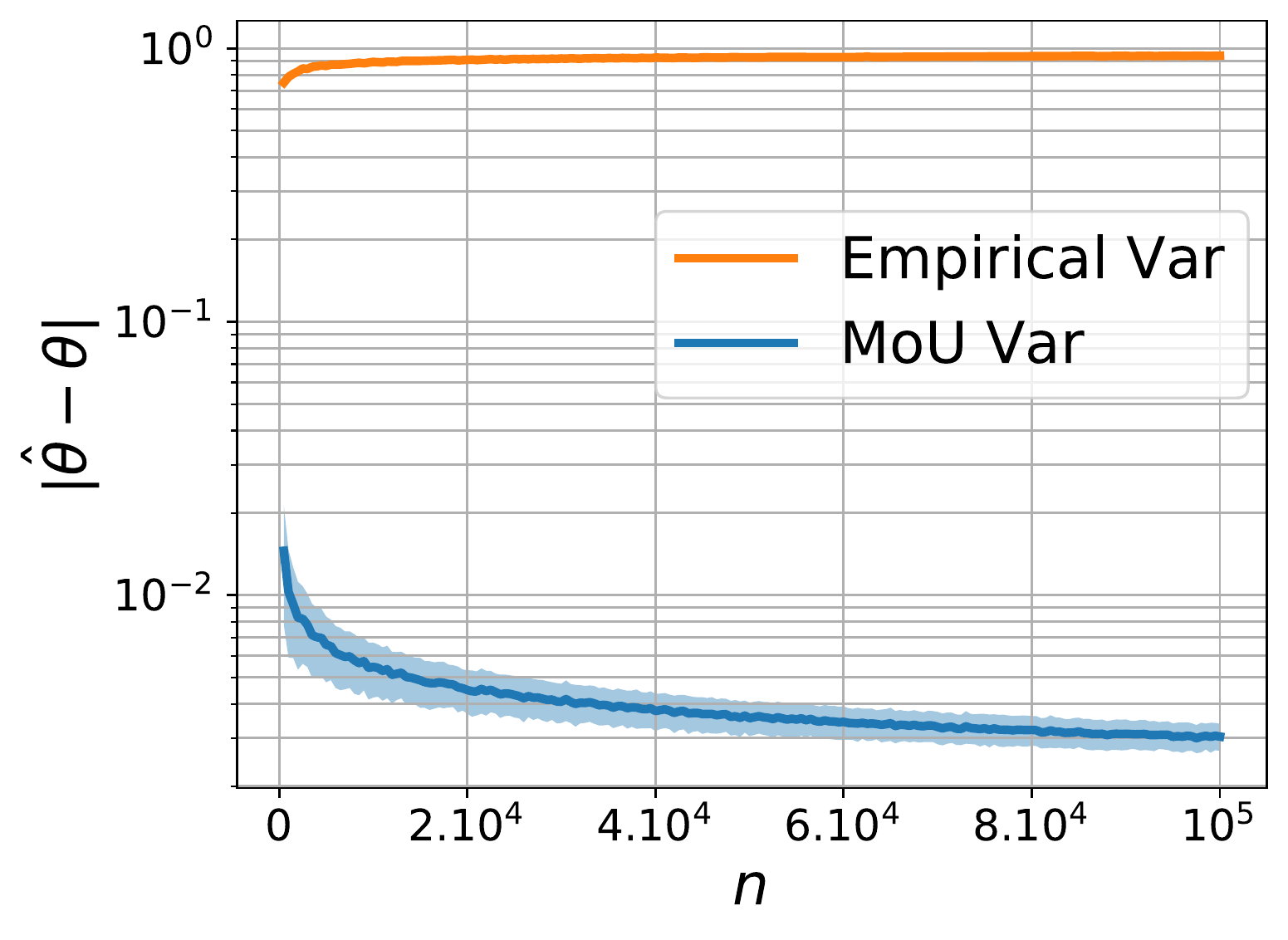}
\caption{MoU estimation of the variance (avg. 500 runs).}
\label{fig:break_var}
\end{subfigure}
\hfill
\begin{subfigure}{0.45\textwidth}
\includegraphics[width=\textwidth]{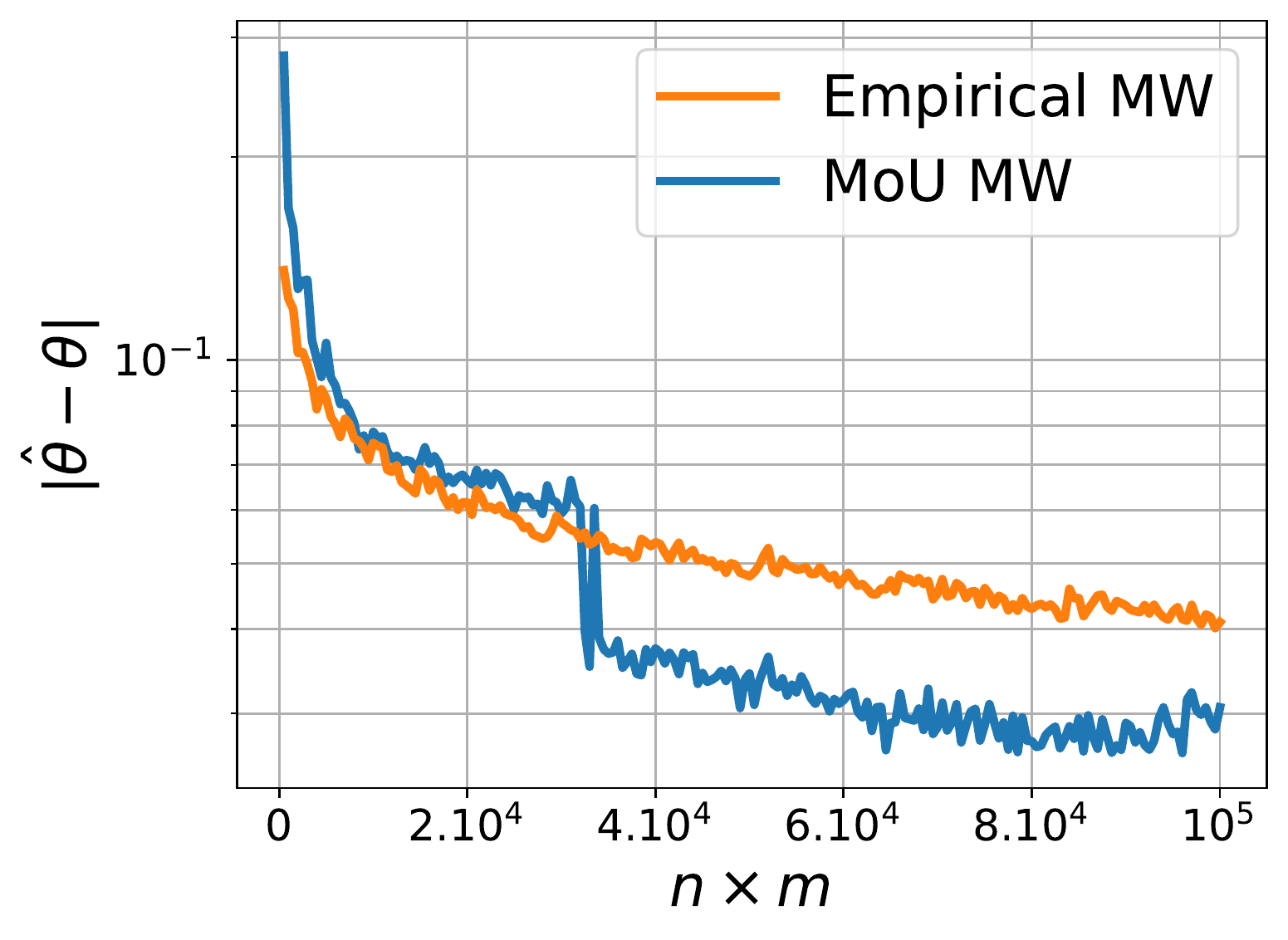}
\caption{MoU estimation of MW-stat (avg. 500 runs).}
\label{fig:break_MW}
\end{subfigure}
\end{center}
\caption{Performances of MoM-based estimators in presence of outliers.}
\end{figure}

{\bf Ruining the mean.}
In this first example, the sane data is drawn according to a standard Gaussian distribution (hence $\theta=0$, and the sub-Gaussian assumption is satisfied with $\rho=1$), and outliers follow a Dirac $\delta_{n^{1/2}}$.
The expected value of the empirical mean estimator $\hat{\theta}_\text{avg}$ is then given by: $\mathbb{E}_{\mathcal{S}_n}[\hat{\theta}_\text{avg}] = (1 - \tau) \cdot 0 + \tau \cdot \sqrt{n} = 1$, always missing the true value.
In contrast, MoM's performance improves with $n$, showing almost no perturbation due to the outliers, see \Cref{fig:break_mean}.

{\bf Ruining the median.}
The Median-of-Means can be seen as an interpolation between the empirical mean (achieved for $K=1$) and the empirical median ($K=n$).
If the first one is known to be very sensitive to abnormal observations, the second is however very robust.
Yet, there are some cases where the median fails and MoM succeeds.
Of course, MoM is a mean estimator while the empirical median estimates the $1/2$ quantile $q_{1/2}$.
Hence, we need to consider a case where both coincide to ensure a fair comparison.
In our second example, sane data follow a Bernoulli of parameter $\theta = 1/2$, and outliers a Dirac $\delta_1$.
When applying blindly the median, one is actually estimating $q_{1/2 + \tau} = 1$.
The results are reported in \Cref{fig:break_median}.
This phenomenon highlights the importance of correctly choosing $\alpha$, a too rough approximation such as the median's leading to poor results.

{\bf Trimmed mean.}
One may argue that a fairer comparison should include the trimmed mean.
However, the latter needs a threshold to be defined, which is hard to set on the basis of the proportion of outliers only.
In contrast, MoM enjoys a closed form formula, depending exclusively on $\tau$, to select the number of blocks $K$ (see \Cref{prop:mom}), that allows to nicely adapt to any contaminated scenario.

{\bf Ruining the variance.}
The empirical variance $\hat{\sigma}^2_n = 1/(n(n-1))\sum_{i < j}(Z_i-Z_j)^2$ is a typical example of a (1-sample) $U$-statistic of degree $2$, with kernel $h\colon (Z, Z') \mapsto (Z - Z')^2/2$.
Our third setting is as follows: $Z$ follows a uniform law on $[0, 1]$ (so that $\theta = 1/12$, and the supremum of $h(Z, Z')$ is finite equal to $1/2$), while outliers are drawn according to the Dirac $\delta_{n^{1/4}}$.
Similarly to the mean, one then has $\mathbb{E}_{\mathcal{S}_n}\left[\hat{\sigma}^2_n\right]$ of the order of $1$, no matter the number of observations considered.
In contrast, MoU behaves almost as if the dataset were not contaminated, see \Cref{fig:break_var}.

{\bf Estimating the Mann-Whitney statistic.}
A classical $2$-sample $U$-statistic of degrees $(1, 1)$ is the Mann-Whitney statistic.
Given two random variables $X$ and $Y$, it aims at estimating $\mathbb{P}\left\{X \le Y\right\}$.
From two samples of realizations $(X_1, \ldots X_n)$ and $(Y_1, \ldots, Y_m)$ of $X$ and $Y$, it is computed~by: $\hat{U}^\text{MW}_{n, m} = 1/(nm) \sum_{i=1}^n\sum_{j=1}^m \mathbbm{1}\{X_i \le Y_j\}$.
This example is very interesting as it highlights the importance of the bounded assumption.
Indeed, to get the convergence of MoU$_2$, we only need boundedness of $H$ on the inliers.
In particular, examples $a)$ and $c)$ above use the unboundedness of the kernel on the outliers to make the empirical mean (respectively variance) arbitrary far away from the true value.
Here, since the kernel $H\colon(X,Y) \mapsto \mathbbm{1}\{X \le Y\}$ is always bounded, the empirical version actually shows more resistance, and the advantage of MoU$_2$ is less important than in other configurations, see \Cref{fig:break_MW}.

\subsection{Learning Experiments}
\label{exp:learning}

Learning experiments have been run in order to highlight the good generalization capacity of MoU minimizers, theoretically established in \Cref{thm:mou_min,thm:alg}.
We considered two pairwise learning problems, \emph{metric learning} and \emph{ranking}, on three benchmark datasets (\emph{iris}, \emph{boston housing} and \emph{wine quality}).
We first corrupted the datasets, in a way described below, before running \Cref{alg:gd}.

\begin{figure}[!t]
\begin{center}
\begin{subfigure}{0.47\textwidth}
\includegraphics[width=\textwidth]{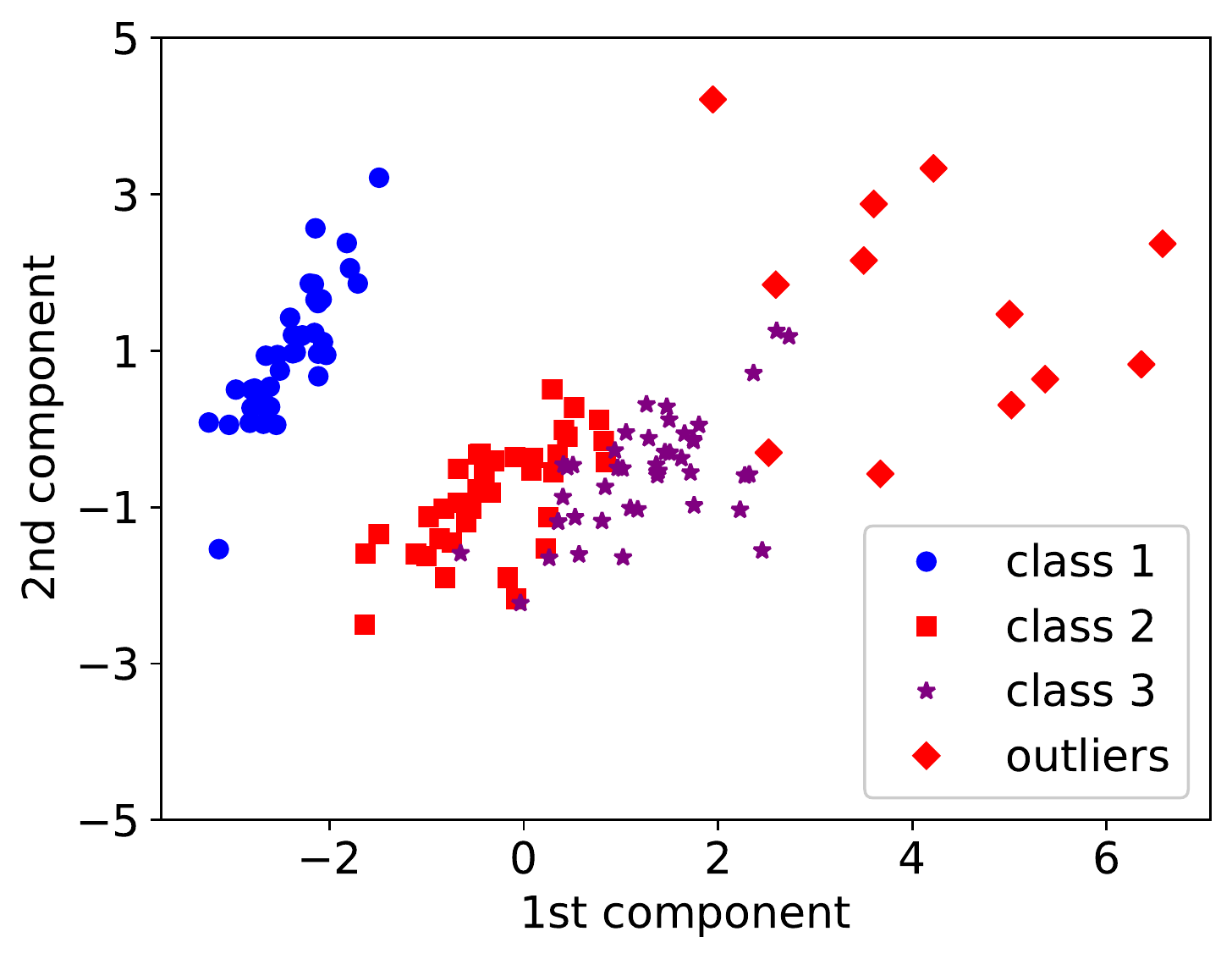}
\caption{Contamination of the \emph{iris} dataset}
\label{fig:iris_cont}
\end{subfigure}
\hfill
\begin{subfigure}{0.47\textwidth}
\vspace{1.1cm}
\begin{center}
\begin{tabular}{cccc}\toprule
                               &       & {\sc GD} & {\sc MoU-GD}\\\midrule
\multirow{2}{*}{\emph{boston}} & sane  & \textbf{0.35}~$\pm$~\textbf{0.04} & $0.36\pm0.05$\\
                               & cont. & 0.99~$\pm$~0.68                   & \textbf{0.36}~$\pm$~\textbf{0.05}\\\midrule
\multirow{2}{*}{\emph{wine}}   & sane  & \textbf{0.73}~$\pm$~\textbf{0.02} & 0.74~$\pm$~0.02\\
                               & cont. & 0.92~$\pm$~0.11                   & \textbf{0.74}~$\pm$~\textbf{0.02}\\\bottomrule
\end{tabular}
\vspace{1.4cm}
\caption{Ranking test losses (avg. 50 runs).}
\label{tab:ranking}
\end{center}
\end{subfigure}\\[0.5cm]
\begin{subfigure}{0.47\textwidth}
\includegraphics[width=\textwidth]{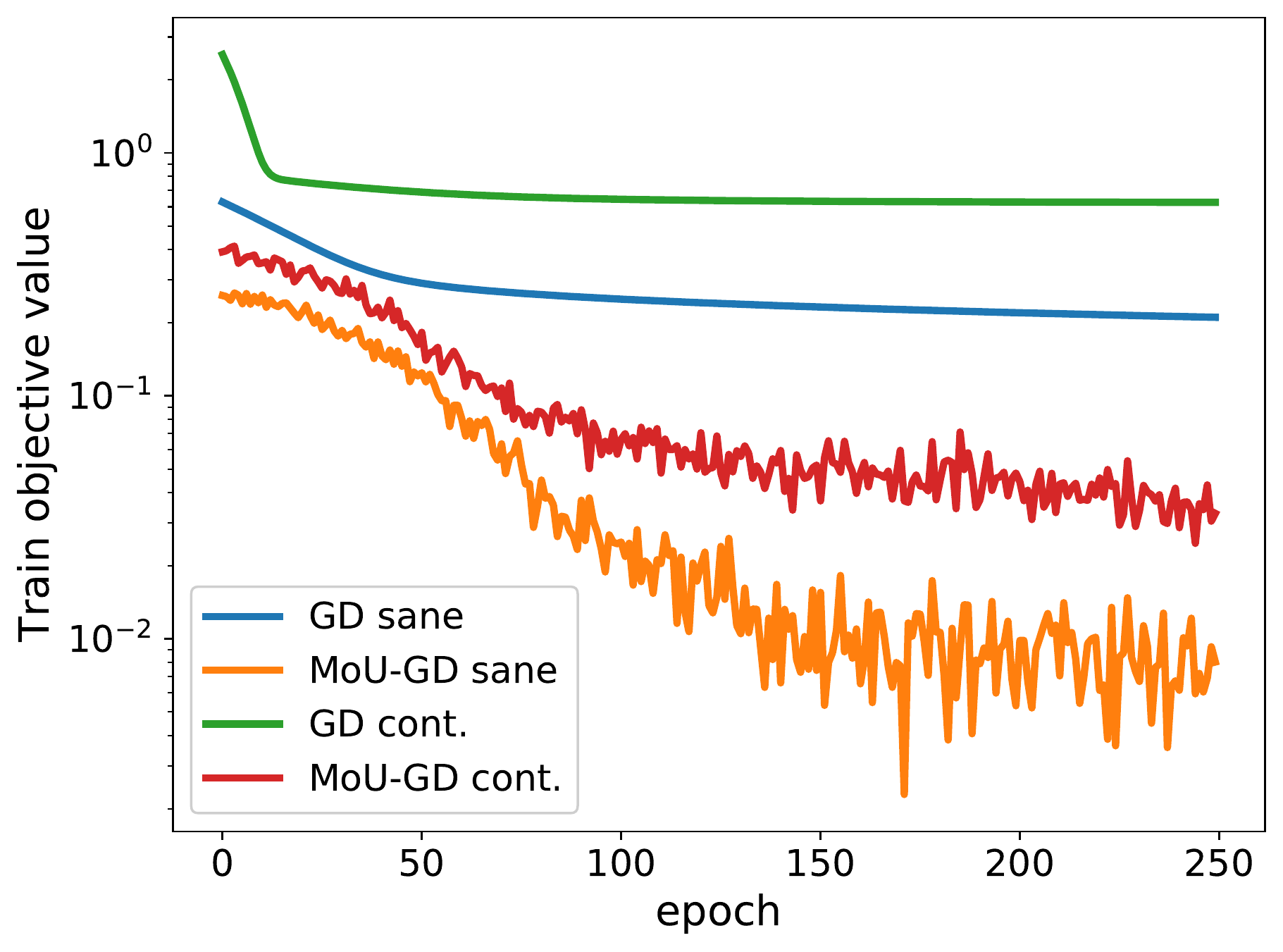}
\caption{Gradient descents on train dataset (avg. 100 runs).}
\label{fig:iris_train}
\end{subfigure}
\hfill
\begin{subfigure}{0.47\textwidth}
\includegraphics[width=\textwidth]{Expes/res/iris_test}
\caption{Gradient descents on test dataset (avg. 100 runs).}
\label{fig:iris_test}
\end{subfigure}
\end{center}
\caption{Performances of MoU-Gradient Descent.}
\end{figure}

{\bf Metric Learning.}
In metric learning, one is interested in learning a distance $d\colon \mathcal{X} \times \mathcal{X} \rightarrow \mathbb{R}_+$, that coincides with some \emph{a priori} information.
We considered the set of Malahanobis distances on $\mathbb{R}^q$ $d^2_M\colon (x, x') \mapsto (x - x')^\top M (x - x')$, with $M \in \mathbb{R}^{q \times q}$ positive semi-definite, and the \emph{iris} dataset\footnote{\href{https://scikit-learn.org/stable/modules/generated/sklearn.datasets.load_iris.html}{https://scikit-learn.org/stable/modules/generated/sklearn.datasets.load\_iris.html}}, that gathers $4$ attributes (sepal length, sepal width, petal length, and petal width) of $150$ flowers issued from $3$ different types of irises.
The \emph{a priori} information we want our distance to match is the class, as we want flowers coming from the same class to be close according to our metric, and conversely.
Denoting $y_{ij} = 2\cdot\mathbbm{1}\{y_i=y_j\} - 1$, the (pairwise) criterion we want to optimize writes as follows:
\begin{equation*}
\min_{M \in S_q^+(\mathbb{R})} ~~ \frac{2}{n(n-1)} \sum_{i < j} \max\Big(0, 1 + y_{ij}(d^2_M(x_i, x_j) - 2)\Big).
\end{equation*}
The whole dataset is first normalized and divided into a train set of size $80\%$ and a test set of size $20\%$.
Then, the training data is contaminated with $10\%$ of outliers drawn uniformly over $[0, 5]^4$, and with label $2$, see \Cref{fig:iris_cont}.
Standard and MoU Gradient Descents are run (with a projection step on $S_q^+(\mathbb{R})$, and $K$ chosen according to the harmonic upper bound), on both the contaminated dataset and the original one of size $80\%$.
The trajectories of the descents averaged over $100$ runs are plotted in \Cref{fig:iris_train} for the train objective, and in \Cref{fig:iris_test} for the test one.
MoU-GD remarkably resists to the presence of outliers, and shows test performance comparable to the sane GD.
In contrast, the contaminated GD converges towards a completely shifted parameter, degrading dramatically its test performance.
The erratic convergence of MoU-GDs is due to the fact that the objective monitored is the sum of distances on the median block only, that is shuffled at each iteration.
This also explains their lower values.
The fact that MoU-GD performs better on the contaminated dataset might not be so surprising.
MoM-based approaches discard data.
When the latter is not relevant or contaminated, this is an undeniable advantage.
When all data are informative, keeping the median block discards the more discriminative points, explaining the slower convergence.
Notice furthermore that MoU-GD on the sane dataset has been run with a value of $K$ designed for the contaminated one.
Strictly following the Harmonic upper bound one should have chosen instead $K = 1$ (since $\tau=0$), and would have recovered the standard GD.
However, since in practice the proportion of outliers is generally unknown, it appeared reasonable to apply the same $K$.
This indeed provides as very interesting tradeoff: it does not affect too much the convergence if the dataset is sane, and prevents from diverging if outliers are present.
The code used is in Python, and has the same computational complexity as the standard Gradient Descent.
It is attached with the submission for reproducibility purpose.

{\bf Ranking.}
In ranking, the observations available to the practitioner are typically composed of feature vectors $X \in \mathbb{R}^p$ describing different objects, and labels $Y \in \mathbb{R}$ representing how much the objects are appreciated by some subject.
One is then interested in learning a decision rule $g\colon \mathbb{R}^p \times \mathbb{R}^p \rightarrow \{-1, 1\}$ to predict if object $X$ is preferred over object $X'$ (\textit{i.e.} $Y \ge Y'$).
We considered the set of decision functions deriving from a scoring function $s\colon \mathbb{R}^p \rightarrow [0, 1]$ such that $g(X, X') = 2\cdot\mathbbm{1}\{s(X) \ge s(X')\} - 1$.
The scoring functions themselves are indexed by vectors $w \in \mathbb{R}^p$ such that $s(x) = \sigma(w^\top x)$, with $\sigma$ the sigmoid function.
ERM then consists in minimizing the disagreements among the training pairs, that writes:
\begin{equation*}
\min_{w \in \mathbb{R}^p} ~~ \frac{2}{n(n-1)}\sum_{i < j}\mathbbm{1}\{g_w(X, X')(Y - Y') \le 0\},
\end{equation*}
and can be relaxed into:
\begin{equation}\label{eq:crit}
\min_{w \in \mathbb{R}^p} ~~ \frac{2}{n(n-1)}\sum_{i < j}\max\Big(0, 1 - g_w(X, X')(Y - Y')\Big).
\end{equation}
We have run \Cref{alg:gd} with criterion \eqref{eq:crit} on two datasets: \emph{boston housing}\footnote{\href{https://scikit-learn.org/stable/modules/generated/sklearn.datasets.load_boston.html}{https://scikit-learn.org/stable/modules/generated/sklearn.datasets.load\_boston.html}}, that gathers $506$ houses described by $13$ real features (\textit{e.g.} number of rooms, distance to employment centers), along with a label corresponding to their prices (real, between $5$ and $50$), and \emph{red wine quality}\footnote{\href{https://archive.ics.uci.edu/ml/datasets/wine+quality}{https://archive.ics.uci.edu/ml/datasets/wine+quality}}, that gathers $1,600$ wines described by $12$ chemical features, along with a label corresponding to a note between $0$ and $10$.
The datasets have first been normalized, and divided into a train set of size $80\%$, and a test set of size $20\%$.
The outliers have then been generated as follows.
A standard GD is first run on the sane training dataset, returning an optimal vector $\hat{w}_\text{sane}$.
Then, $2\%$ and $5\%$ of outliers (for \emph{boston} and \emph{wine} respectively) have been generated by sampling $(X_\text{outlier}, Y_\text{outlier})$ uniformly around $(-\lambda \hat{w}_\text{sane}, \lambda)$, for some real value $\lambda$.
This way, one has:
\begin{align*}
g_{\hat{w}_\text{sane}}(X, X_\text{outlier})(Y - Y_\text{outlier}) &\approx \left(\sigma(\hat{w}_\text{sane}^\top X) - \sigma(\hat{w}_\text{sane}^\top X_\text{outlier})\right)(Y - \lambda),\\[0.2cm]
&= \left(\sigma(\hat{w}_\text{sane}^\top X) - \sigma(-\lambda\|\hat{w}_\text{sane}\|^2)\right)(Y - \lambda).
\end{align*}

Making $\lambda$ tend to $+\infty$ (respectively $-\infty$), the first term becomes always positive and the second very negative (respectively always negative and very positive), incurring important losses preventing from converging toward $\hat{w}_\text{sane}$.
For \emph{boston}, $\lambda$ was set to $-500$, and to $50$ for \emph{wine}.
The GD trajectories obtained are very similar to that of the metric learning example, and are thus not reproduced here.
The generalization errors obtained on the test dataset of size $20\%$ are gathered in Table~\ref{tab:ranking}.
Again, MoU-GD shows a remarkable resistance to the presence of outliers, and attains almost the same performance as standard GD on the sane dataset.
This little gap may be partly due to the instability of MoU-GD (see \textit{e.g.} \Cref{fig:iris_train}), which uses mini-batches.

\end{document}